\documentclass{article}
\usepackage{main_preprint,times}

\usepackage{amsmath,amsfonts,bm}

\def\eqref#1{equation~\ref{#1}}

\def\1{\bm{1}}

\DeclareMathAlphabet{\mathsfit}{\encodingdefault}{\sfdefault}{m}{sl}
\SetMathAlphabet{\mathsfit}{bold}{\encodingdefault}{\sfdefault}{bx}{n}

\usepackage{hyperref}
\usepackage{url}
\usepackage{latexsym}
\usepackage[T1]{fontenc}
\usepackage[utf8]{inputenc}
\usepackage{microtype}
\usepackage{inconsolata}
\usepackage{graphicx}
\usepackage{booktabs}
\usepackage{multicol}
\usepackage{amsmath}
\usepackage{amsfonts}
\usepackage{multirow}
\usepackage{enumitem}
\usepackage{wrapfig}
\usepackage[table]{xcolor}
\usepackage{xcolor}
\usepackage{tcolorbox}
\newtcolorbox{takeaway}{colback=lightgray!10, colframe=black, boxrule=0.4pt, arc=1pt,
  left=3pt, right=3pt, top=1pt, bottom=1pt, boxsep=1pt,
  before skip=4pt, after skip=4pt}
\usepackage{subcaption}
\newcommand{\huggingfacedown}{\includegraphics[height=0.75em]{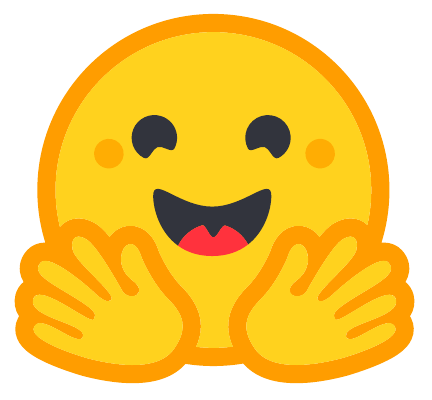}}
\newcommand{\githubdown}{\includegraphics[height=0.75em]{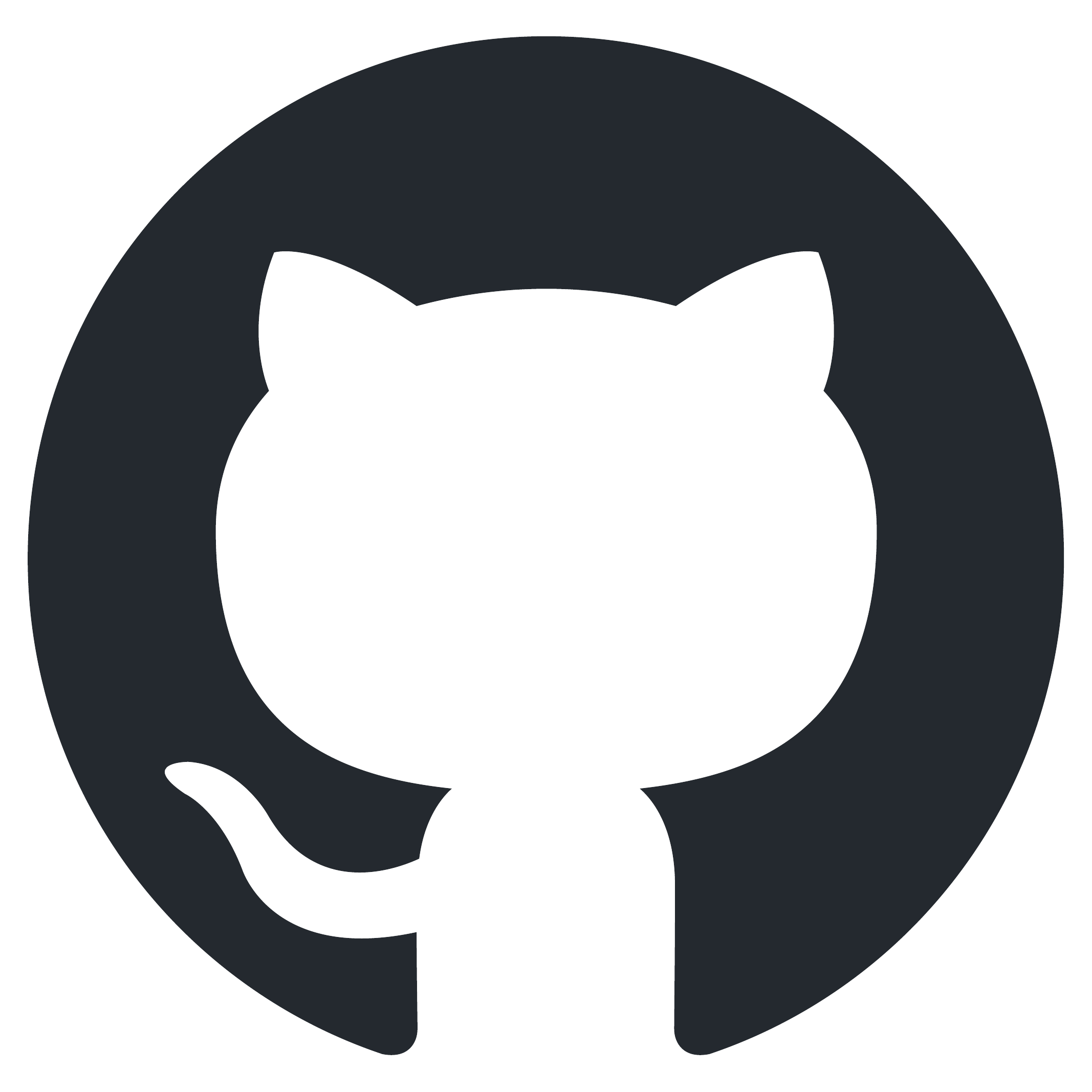}}

\title{Synthetic Pre-pretraining Survives Scale, but Not as a Grammatical Prior}

\author{Atsuki Yamaguchi{\rm \textsuperscript{1}}
\quad
Tatsuro Inaba{\rm \textsuperscript{2}}
\quad
Joel Niklaus{\rm \textsuperscript{3}}
\quad
Michal Štefánik{\rm \textsuperscript{4}}\\
\textbf{Aline Villavicencio}\textsuperscript{1,5,6}
\quad
\textbf{Nikolaos Aletras}\textsuperscript{1}\\\\
\textsuperscript{1}University of Sheffield, UK
\quad\textsuperscript{2}Mohamed bin Zayed University of Artificial Intelligence, UAE\\
\textsuperscript{3}Hugging Face
\quad\textsuperscript{4}National Institute of Informatics, Japan
\quad\textsuperscript{5}University of Exeter, UK\\
\textsuperscript{6}Federal University of Rio Grande do Norte, Brazil
}

\iclrfinalcopy
\begin{document}

\maketitle

\begin{abstract}
Pre-pretraining (PPT) on synthetic non-natural language data improves token efficiency during language model pre-training (PT).
Prior work attributes this gain to a grammatical prior, i.e., a structural inductive bias learned during PPT that transfers to natural language grammar.
However, PPT has only been tested on models of at most 1B parameters and PT budgets below 2B tokens on predominantly web text. It is unknown whether PPT is effective at larger scales and under more realistic PT data mixtures that combine diverse sources (e.g., code and math).
We therefore present a comprehensive study on PPT spanning five PPT tasks, four PT data mixtures, four parameter scales (500M to 7B), and PT budgets of up to 100B tokens.
Our results demonstrate that the downstream performance and token efficiency gains of PPT persist at scale, e.g., saving at least 21B PT tokens at the 3B scale.
However, in contrast to prior work, we find no consistent evidence that these gains stem from a grammatical prior.
Downstream performance does not consistently align with grammatical acceptability across model sizes.
Instead, we find that downstream gains arise from PPT tasks that improve long-range retrieval.
Finally, PPT performance gains are robust to how PT data mixtures are composed and diminish only when web text is absent.
Overall, PPT is a low-cost addition to PT, and future PPT task design should target long-range retrieval rather than natural language grammar.
\end{abstract}

\vspace{-1em}
\quad\quad\quad\quad\githubdown\quad\url{https://github.com/gucci-j/verify-ppt-at-scale}\\

\vspace{-1.5em}
\quad\quad\quad\quad\huggingfacedown\quad\url{https://huggingface.co/verify-ppt}
\vspace{-0.5em}

\section{Introduction}
\label{sec:intro}
\begin{wrapfigure}{r}{0.38\textwidth}
\vspace{-1em}
\centering
\begin{subfigure}{\linewidth}
  \centering
  \includegraphics[width=\linewidth]{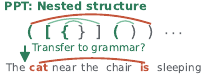}
  \caption{Hypothesis under test.}
  \label{fig:motivation_hypothesis}
\end{subfigure}\\[0.35em]
\begin{subfigure}{\linewidth}
  \centering
  \includegraphics[width=\linewidth]{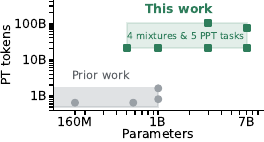}
  \caption{Design space.}
  \label{fig:motivation_design}
\end{subfigure}
\caption{\textbf{(a)} PPT is claimed to induce a grammatical prior that transfers to natural language.
\textbf{(b)} Prior work sits at or below 1B parameters and short training horizons. We span four scales, four PT mixtures, five PPT tasks, and up to 100B PT tokens.}
\label{fig:motivation}
\vspace{-2.25em}
\end{wrapfigure}

Pre-training (PT) equips language models (LMs) with general capabilities~\citep[\textit{inter alia}]{gemmateam2026gemma4technicalreport,qwenteam2026qwen35omnitechnicalreport,kimiteam2026kimik3openfrontier} but is prohibitively expensive, often consuming tens of trillions of tokens~\citep{NEURIPS2022_c1e2faff}.
Recent studies~\citep[\textit{inter alia}]{hu-etal-2025-circuits,mita-etal-2026-language,cheng2026logiclanguageprepretrainingformal} show that pre-pretraining (PPT), a warm-up phase on synthetic non-natural language sequences such as $k$-Shuffle Dyck (i.e., an interleaved balanced-parentheses language), improves token efficiency during subsequent PT on natural language data.
Previous work~\citep{hu-etal-2025-circuits,mita-etal-2026-language} attributes this improvement to a \textit{grammatical prior}, i.e., a structural inductive bias acquired from PPT data, which transfers to natural language (Figure~\ref{fig:motivation}a).

The practical utility of PPT rests on one question that remains unanswered under more realistic conditions of larger model size, longer PT, and more diverse PT data: \textit{do the performance and token efficiency gains of PPT persist at scale?}
Prior studies on PPT cannot answer this question, because their experimental setups diverge from typical PT scenarios in three ways (Figure~\ref{fig:motivation}b).
First, they experiment with models at or below 1B parameters~\citep{hu-etal-2025-circuits,mita-etal-2026-language,jiang2026procedural,guo2026syntheticprepretrainingimproveslanguage,cheng2026logiclanguageprepretrainingformal}, whereas PT studies often operate at 2B parameters and above to draw conclusions~\citep{pmlr-v202-biderman23a,NEURIPS2023_fa3ed726,xue2026multihashformerhashbasedgenerativelanguage}.
\citet{hu-etal-2025-circuits} and \citet{mita-etal-2026-language} leave the behavior of PPT beyond 1B parameters as an open question.
Second, they often stop PT early at less than 2B tokens~\citep[\textit{inter alia}]{hu-etal-2025-circuits,mita-etal-2026-language}, far short of the budgets used to validate PT design decisions in recent work~\citep[e.g., 100B tokens]{cheng-etal-2024-instruction,ye2025data,yamaguchi-etal-2026-enhancing}.
This raises the possibility that observed gains stem from temporary initialization effects.
Such benefits may disappear as extended optimization shifts parameters away from the warm-up state~\citep{NEURIPS2020_c76e4b2f,pmlr-v202-liu23ao}.
Third, they rely on single-domain PT corpora, e.g., C4~\citep{c4}, that are dominated by or entirely derived from web text~\citep{hu-etal-2025-circuits,mita-etal-2026-language}.
Modern PT instead employs curated mixtures containing code and mathematics data~\citep{bakouch2025smollm3,martins2025eurollm9btechnicalreport,olmo2026olmo3}.
Because code and mathematics contain nested dependencies and strict structural rules, these domains may supply structural signals similar to those from PPT, potentially rendering it redundant.\looseness=-1

In this paper, we systematically evaluate PPT across four parameter scales (500M, 1B, 3B, and 7B) under a default PT budget of 21B tokens and an extended budget reaching 100B tokens, i.e., up to around 50$\times$ more than previous work.
We test four publicly available PT data mixtures with documented composition: web text (C4); a web-dominant mixture (Marin\footnote{\url{https://marin.readthedocs.io/en/latest/reports/marin-8b-retro/}}); a filtered-web mixture with balanced code and mathematics (SmolLM3~\citep{bakouch2025smollm3}); and a STEM-weighted mixture (OLMo3~\citep{olmo2026olmo3}).
We also compare five PPT tasks: two formal languages; two structured synthetic tasks lacking formal grammars; and an in-domain text control that samples PPT data from the PT corpus.
We evaluate each model on linguistic competence via BLiMP~\citep{warstadt-etal-2020-blimp-benchmark} for grammatical acceptability and verbatim retrieval~\citep{armeni-etal-2022-characterizing,armeni-etal-2024-transformer} for long-range retrieval capability.
Furthermore, we test general capability across ten benchmarks spanning reading comprehension, science question answering (QA), commonsense reasoning, and language modeling.

Our key contributions are as follows:
\begin{itemize}[leftmargin=*, itemsep=0.5em, topsep=0pt, partopsep=0pt, parsep=0pt]
\item We present the first systematic evaluation of PPT at model and data scale, spanning 500M to 7B model parameters, PT budgets up to 100B tokens, four PT data mixtures, and five PPT tasks.
\end{itemize}

\vspace{-1.4pt} %
\begin{itemize}[leftmargin=*, itemsep=0.5em, topsep=0pt, partopsep=0pt, parsep=0pt]
\item We show that the downstream performance and token efficiency benefits of PPT persist under parameter scaling and extended PT up to 100B tokens, saving at least 21B PT tokens at the 3B scale, and remain positive even at 7B.

\item However, we find no consistent evidence that considering PPT as a grammatical prior~\citep{hu-etal-2025-circuits} explains the effectiveness of PPT.
Downstream performance does not consistently align with grammatical acceptability.
Instead, we find that downstream gains arise only from PPT tasks that improve long-range retrieval.

\item We show that PPT gains depend on the presence of web text rather than the proportion of code and mathematics. A 13$\times$ increase in mathematical data leaves the downstream gain from PPT unchanged, whereas removing web text collapses it.
\end{itemize}

\section{Related Work}
\subsection{Synthetic Pre-pretraining}
\label{subsec:rw_synthetic}

Early work has explored structural transfer by pre-training LSTMs~\citep{10.1162/neco.1997.9.8.1735} and Transformers~\citep{NIPS2017_3f5ee243} on artificial languages or non-linguistic sequences, including music~\citep{papadimitriou-jurafsky-2020-learning} and artificial formal grammars~\citep{papadimitriou-jurafsky-2020-learning,ri-tsuruoka-2022-pretraining,papadimitriou-jurafsky-2023-injecting}, prior to fine-tuning or evaluating them on natural language tasks.
In modern decoder-only LMs, PPT is a warm-up training phase on synthetic sequences before PT on natural language data, consisting of three steps.
First, a generator produces a synthetic corpus, typically from a formal language such as $k$-Shuffle Dyck~\citep{hu-etal-2025-circuits}.
Second, a randomly initialized model trains on the generated corpus for a budget far smaller than the subsequent PT run.
Third, the resulting parameters initialize PT, either transferred in full~\citep{hu-etal-2025-circuits,mita-etal-2026-language} or with the embedding matrix and LM head re-initialized for the natural language tokenizer~\citep{cheng2026logiclanguageprepretrainingformal,lee2026traininglanguagemodelsneural}.

Prior work on PPT differs mainly in what underlying structure the synthetic corpus encodes.
\citet{hu-etal-2025-circuits} argue that an effective corpus should capture hierarchical dependencies while remaining learnable by a Transformer. They instantiate this with $k$-Shuffle Dyck, a context-sensitive bracket language to generate synthetic PPT data.
\citet{mita-etal-2026-language} claim that bracket matching in $k$-Shuffle Dyck lacks cues to which opening bracket a closing one resolves. To tackle this issue, they propose adding agreement and displacement to the bracketed structure.
Beyond formal grammars, PPT has expanded to algorithmic tasks~\citep{jiang2026procedural}, formal logical derivations~\citep{cheng2026logiclanguageprepretrainingformal}, state trajectories of neural cellular automata~\citep{lee2026traininglanguagemodelsneural}, and synthetic recurrent structures~\citep{guo2026syntheticprepretrainingimproveslanguage}, with recent extensions reaching vision models~\citep{Shinnick_2026_CVPR}.

\subsection{Interplay between Pre-pretraining and Pre-training}

\citet{hu-etal-2025-circuits} attribute PPT effectiveness to a grammatical prior.
Models acquire structural priors independently of natural language vocabulary or world knowledge, and these priors transfer to natural language grammar~\citep{papadimitriou-jurafsky-2020-learning,ri-tsuruoka-2022-pretraining,papadimitriou-jurafsky-2023-injecting}.
However, this explanation has been tested only on predominantly web-text data, leaving unexamined how PPT interacts with PT data mixtures that combine more diverse sources, e.g., code and mathematics.
Such PT mixtures may induce structural abstractions~\citep{kim2024codepretrainingimprovesentity,petty2025how}, making PPT unnecessary.

\subsection{Scale, Optimization Dynamics, and Warm-up Retention}
\label{subsec:rw_scale}

Three conditions related to the experimental settings used in PPT challenge the generalizability of earlier findings.
First, prior evaluations generally operate at or below the 1B parameter scale~\citep{hu-etal-2025-circuits,budnikov-yamshchikov-2025-transfer,mita-etal-2026-language,jiang2026procedural,guo2026syntheticprepretrainingimproveslanguage,cheng2026logiclanguageprepretrainingformal}.
Larger Transformers, however, can internalize hierarchical abstractions through standard training objectives alone~\citep{pmlr-v202-liu23ao,allen-zhu2025physics}.
Therefore, larger model capacity may already supply what synthetic PPT provides.
Second, previous studies rely on small batch sizes (e.g., 32) in both PPT and PT~\citep{hu-etal-2025-circuits,mita-etal-2026-language,guo2026syntheticprepretrainingimproveslanguage}, introducing higher stochastic gradient noise than large-batch setups~\citep{mccandlish2018empiricalmodellargebatchtraining,l.2018dont}.
In such noisy regimes, reported advantages may reflect initialization variance rather than a durable bias.
Third, existing work often stops PT early ($\le$ 2B tokens)~\citep{hu-etal-2025-circuits,budnikov-yamshchikov-2025-transfer,mita-etal-2026-language,guo2026syntheticprepretrainingimproveslanguage}, whereas extended optimization may dilute initialization biases that persist in under-trained models~\citep{NEURIPS2020_c76e4b2f,pmlr-v202-liu23ao}.\looseness=-1

\section{A Systematic Framework for Evaluating Pre-pretraining}

\subsection{Problem Setting}
\label{subsec:problem_setting}

Let $\mathcal{M}_\theta$ be an autoregressive LM with weights $\theta \in \mathbb{R}^N$, where $N$ is the number of parameters, and let $\mathcal{L}(\theta; \mathcal{D}) = \mathbb{E}_{x \sim \mathcal{D}} \left[ -\sum\nolimits_{t=1}^{|x|} \log p_\theta(x_t \mid x_{<t}) \right]$ denote the negative log-likelihood (NLL) over a corpus $\mathcal{D}$.
Standard PT minimizes $\mathcal{L}(\theta; \mathcal{D}_{\text{PT}})$ from a random initialization $\theta_0$, yielding $\theta_{\text{PT}}$ (\textbf{PT-Only}).
PPT first minimizes $\mathcal{L}(\theta; \mathcal{D}_{\text{PPT}})$ from the same $\theta_0$ over a much smaller dataset $\mathcal{D}_{\text{PPT}}$ where $|\mathcal{D}_{\text{PPT}}| \ll |\mathcal{D}_{\text{PT}}|$, yielding $\theta_{\text{PPT}}$.
These weights then replace $\theta_0$ as the initialization state for PT.\looseness=-1

To test whether the benefits of PPT generalize beyond small-scale settings, we evaluate performance across four dimensions: the PPT task (\S\ref{subsec:ppt_tasks}), PT data mixture (\S\ref{subsec:data_mixtures}), parameter scale (\S\ref{subsec:scale}), and PT budget (\S\ref{subsec:duration}).\looseness=-1

\subsection{Pre-pretraining Data}
\label{subsec:ppt_tasks}
We evaluate five approaches for constructing $\mathcal{D}_{\text{PPT}}$: two based on formal languages, two structured synthetic tasks without formal grammars, and an in-domain text control where we sample $\mathcal{D}_{\text{PPT}}$ data from $\mathcal{D}_{\text{PT}}$.
This allows us to test whether transfer requires a formal grammar or follows from structured sequences of any kind.

\paragraph{Formal Languages.}
\textbf{$k$-Shuffle Dyck}~\citep{hu-etal-2025-circuits} is a context-sensitive language of interleaved bracket pairs (e.g., \texttt{( [ ( ] ) )}).
As the foundational PPT task for which transfer to natural language grammar was reported, it serves as our primary synthetic task.
\textbf{MP-Struct Core}~\citep{mita-etal-2026-language} places marker tokens beside each bracket pair (e.g., \texttt{[0 H\_C (4 )4 ]0}), making the matching bracket unambiguous, whereas $k$-Shuffle Dyck leaves several candidates open (e.g., the first `\texttt{)}' above has two open `\texttt{(}' candidates).

\paragraph{Structured Synthetic Tasks without Formal Grammars.}
\textbf{Set} \citep{jiang2026procedural} removes duplicate tokens while preserving their original order (e.g., \texttt{1 2 2 | 1 2}). 
This requires tracking previously seen tokens, but no hierarchical recursion.
The \textbf{neural cellular automata} (\textbf{NCA}) task~\citep{lee2026traininglanguagemodelsneural} consists of successive states of an NCA, where a fixed rule updates each cell from its neighbors.
The resulting dependencies repeat over time and are not nested.

\paragraph{In-domain Text Control.}
We introduce a natural-language control task (\textbf{Control}) where we sample sequences for $\mathcal{D}_{\text{PPT}}$ from $\mathcal{D}_{\text{PT}}$, disjoint from samples seen during PT.
This separates the effect of synthetic sequences from that of the extra optimization steps on performance.

\subsection{Pre-training Data Mixture}
\label{subsec:data_mixtures}
\begin{wraptable}{r}{0.41\textwidth}
\begin{center}
\small
\vspace{-3.5em}
\renewcommand{\arraystretch}{1.0}
\setlength{\tabcolsep}{3pt}
\caption{Comparison of PT data mixtures, as reported by their creators.
Web is a subset of natural language, with the remainder drawn from non-web sources such as books, academic text, and encyclopedic text.
Our PT runs follow these ratios.
}
\label{tab:data}
\resizebox{\linewidth}{!}{
\begin{tabular}{lcccc}
\toprule
\textbf{Category (\%)} & \textbf{C4} & \textbf{SmolLM3} & \textbf{OLMo3} & \textbf{Marin} \\
\midrule
Natural Lang. & 100.0 & 85.0 & 89.5& 92.6\\
\quad {\scriptsize $\llcorner$Web} & {\scriptsize 100.0} & {\scriptsize 79.8} & {\scriptsize 76.9} & {\scriptsize 92.6}\\ \midrule
Code & 0.0 & 12.0 & 7.1 & 6.1\\
Math & 0.0 & 3.0 & 3.4 & 1.3\\
\bottomrule
\end{tabular}
}
\end{center}
\vspace{-2em}
\end{wraptable}

The composition of $\mathcal{D}_{\text{PT}}$ is itself a variable in modern LM PT, comprising curated mixtures of domains, $\mathcal{D}_{\text{PT}} = \alpha \mathcal{D}_{\text{NL}} + \beta \mathcal{D}_{\text{Code}} + \gamma \mathcal{D}_{\text{Math}}$, where $\alpha + \beta + \gamma = 1$ are the mixing coefficients for natural language, source code, and mathematical data.
We compare four PT data mixtures with varying degrees of code and mathematics to test the interplay between PPT and PT data (Table~\ref{tab:data}).\looseness=-1

\textbf{C4}~\citep{c4} consists entirely of cleaned web text with no code or math ($\beta = \gamma = 0$), where models acquire syntactic knowledge from natural language alone.
Following \citet{hu-etal-2025-circuits}, we adopt it as our reference mixture.
It tests whether earlier PPT approaches survive scaling (\S\ref{subsec:scale} and \S\ref{subsec:duration}).

The other three are curated multi-domain mixtures.
\textbf{SmolLM3} (Stage 1 data)\footnote{Modern PT approaches use a multi-stage curriculum, consisting of a long first stage trained on a broad mixture, followed by shorter stages that upweight curated, high-quality data mixtures. We use Stage 1 mixtures throughout the paper, as these account for the bulk of the PT tokens. Marin refers to this stage as Phase 1.}~\citep{bakouch2025smollm3} pairs heavily filtered web text~\citep{fineweb,NEURIPS2024_19e4ea30,penedo2025fineweb} with the largest combined code and math share, at 15.0\%~\citep{lozhkov2024starcoder2stackv2,allal2025smollm,han-etal-2025-infimm}.
\textbf{OLMo3} (Stage 1)~\citep{olmo2026olmo3} is the most STEM-weighted mixture.
It adds 12.6\% OCR-extracted academic PDFs~\citep{poznanski2025olmocr} to 7.1\% code and 3.4\% math, which leaves the smallest general web share at 76.9\%.
\textbf{Marin} (Phase 1)\footnotemark[2] is web-dominant, combining classifier-filtered DCLM data with light code~\citep{li2023starcoder} and math~\citep{azerbayev2024llemma}.\looseness=-1

\subsection{Model Scales}
\label{subsec:scale}
To evaluate how model capacity interacts with PPT, we test four model scales: 500M, 1B, 3B, and 7B.\looseness=-1

\textbf{500M} and \textbf{1B} scales follow the setting of prior work in model size (\S\ref{subsec:rw_scale}), verifying reported PPT downstream performance and token-efficiency gains.

\textbf{3B} crosses the model size in which \citet{pmlr-v202-biderman23a} report a transition point, where LMs begin to acquire complex factual and structural capabilities that smaller models fail to learn under the same PT data regime.
Theoretical studies~\citep{merrill-etal-2021-effects,strobl-etal-2024-formal} also show that large Transformers can learn hierarchical abstractions directly through standard gradient descent.
This setting therefore tests whether capacity alone makes PPT redundant.

\textbf{7B} extends the model size beyond the $\le$ 1B setting of prior work and is widely used among open-weight releases~\citep{touvron2023llama2openfoundation,qwen2025qwen25technicalreport,k2horizon2026}.
This setting tests whether PPT helps as model capacity grows further.\footnote{Compute constraints limit 7B evaluations to the Marin mixture (up to 75.5B PT tokens) and $k$-Shuffle Dyck.}

\subsection{Pre-training Budget}
\label{subsec:duration}

Prior PPT studies stop PT early, typically after roughly 2B tokens, and train with small batches~\citep{hu-etal-2025-circuits,mita-etal-2026-language,jiang2026procedural,guo2026syntheticprepretrainingimproveslanguage}.
Under these conditions, reported gains may reflect a short-lived initialization advantage or gradient noise rather than a durable bias (\S\ref{subsec:rw_scale}).
We therefore adopt a two-tiered optimization budget.

Our standard PT budget keeps the 10K-step horizon of \citet{hu-etal-2025-circuits} but raises the batch size to 512 sequences (2.1M tokens per step) with a 4,096-token context window, yielding 21B tokens, following a recent approach for effective PT with synthetic data~\citep{niklaus2026how}.
This expansion increases token exposure during training by 12.8 to 25.6$\times$ over prior work~\citep{hu-etal-2025-circuits,mita-etal-2026-language}.
This setup tests whether the PPT benefits persist under large-batch, low-noise optimization with long-context dependencies.

Our extended PT budget reaches 100B tokens (47,684 steps) at 3B, roughly 33 tokens per parameter, which exceeds compute-optimal allocations for multi-billion parameter models~\citep{NEURIPS2022_c1e2faff,chen-etal-2025-revisiting} and matches foundation model recipe ablations~\citep{bakouch2025smollm3,olmo2026olmo3}.
This tests whether the benefit of PPT persists or dilutes over prolonged training.

\section{Experimental Setup}
\subsection{Model Training}
\label{subsec:training_pipeline}

\paragraph{Architecture and Tokenizer.}
All models use the SmolLM3 architecture and tokenizer~\citep{bakouch2025smollm3} with a 128,256-token vocabulary.
We vary only width, depth, and head counts and keep the optimizer, learning rate schedule, batch size, context window, and precision identical (Appendix Table \ref{tab:model_hyperparameters}).
At 7B, we lower the learning rate for training stability while keeping the same schedule.
Performance differences across scales therefore primarily reflect the influence of model capacity.\looseness=-1

\paragraph{Pre-pretraining.}
Each PPT run optimizes $\mathcal{L}(\theta; \mathcal{D}_{\text{PPT}})$ from a random initialization $\theta_0$ for 500 steps, following \citet{hu-etal-2025-circuits}.
As mentioned in \S\ref{subsec:ppt_tasks}, we primarily use $k$-Shuffle Dyck as our PPT task.

\paragraph{Pre-training.}
Following \citet{hu-etal-2025-circuits} and \citet{mita-etal-2026-language}, PPT runs initialize weights from the step-500 checkpoint, $\theta_{\text{PPT}}$, and reset the optimizer moments and scheduler.
A PPT run and its PT-Only counterpart therefore differ only in the initialization state.
See Appendix~\ref{app:training_details} for details.

\subsection{Evaluation}
\label{subsec:evaluation}
\paragraph{Baselines.}
We compare each PPT configuration against two baselines at the same scale and PT data mixture.
\textbf{PT-Only} ($\theta_{\text{PT}}$) trains from a random initialization with no preliminary phase ($\theta_0$), measuring the net effect of PPT.
\textbf{Control} (\S\ref{subsec:ppt_tasks}) serves as a second baseline, as it runs the identical 500-step phase on held-out text from $\mathcal{D}_{\text{PT}}$.

\paragraph{General Capability.}
We evaluate models across ten downstream benchmarks grouped into four categories to test whether PPT translates into downstream performance.
\begin{itemize}[leftmargin=*]
    \item \textbf{Reading Comprehension (RC)}: \textbf{RACE}~\citep{lai-etal-2017-race} and \textbf{ReCoRD}~\citep{zhang2018recordbridginggaphuman}, both zero-shot, scored by accuracy and span F1, respectively.
    
    \item \textbf{Science QA}: zero-shot \textbf{SciQ}~\citep{welbl-etal-2017-crowdsourcing}, plus five-shot \textbf{ARC-Easy}~\citep{clark2018thinksolvedquestionanswering} and \textbf{OpenBookQA}~\citep{mihaylov-etal-2018-suit}, all scored by normalized accuracy.

    \item \textbf{Commonsense Reasoning (CR)}: zero-shot \textbf{HellaSwag}~\citep{zellers-etal-2019-hellaswag} and \textbf{PIQA}~\citep{bisk2020piqa}, scored by normalized accuracy, alongside zero-shot \textbf{COPA}~\citep{gordon-etal-2012-semeval} and five-shot \textbf{SocialIQA}~\citep{sap-etal-2019-social}, scored by accuracy.

    \item \textbf{Language Modeling (LM)}: \textbf{LAMBADA}~\citep{paperno-etal-2016-lambada} (OpenAI version), which requires a final word recoverable only from the full passage, scored by zero-shot accuracy.
\end{itemize}

\paragraph{Linguistic Competence.}
We also examine whether the explanation for the effectiveness of PPT proposed by \citet{hu-etal-2025-circuits}, a grammatical prior evidenced by grammatical acceptability, holds under our expanded scale and PT budgets.
We adopt the same two benchmarks:
(1) \textbf{BLiMP}~\citep{warstadt-etal-2020-blimp-benchmark} measures zero-shot grammatical acceptability over 12 paradigm groups, reporting overall mean accuracy alongside subgroup scores for semantics, morphology, and syntax.
(2) \textbf{Verbatim retrieval}~\citep{armeni-etal-2022-characterizing,armeni-etal-2024-transformer} cues a model to repeat a noun list seen earlier in context, reporting mean NLL over the full stimulus, where lower values indicate more reliable retrieval.

\paragraph{Result Reporting.}
We evaluate each run every 1K steps and report the mean and standard deviation (SD) over the second half of training (i.e., 5K to 10K steps).
We call a PPT gain \textit{stable} when its mean exceeds its SD across these checkpoints.
Downstream conclusions remain invariant to window selection (Appendix~\ref{app:estimator}).
Each configuration is a single run, except for PT-Only and $k$-Shuffle Dyck at 3B on Marin, which we repeat with three random seeds (Appendix~\ref{app:seed}).
Appendix~\ref{app:evaluation_details} gives further evaluation details.

\section{Results}
\label{sec:results}
\subsection{General Capability}
\label{subsec:general}
\begin{figure}[t]
    \centering
    \includegraphics[width=0.99\linewidth]{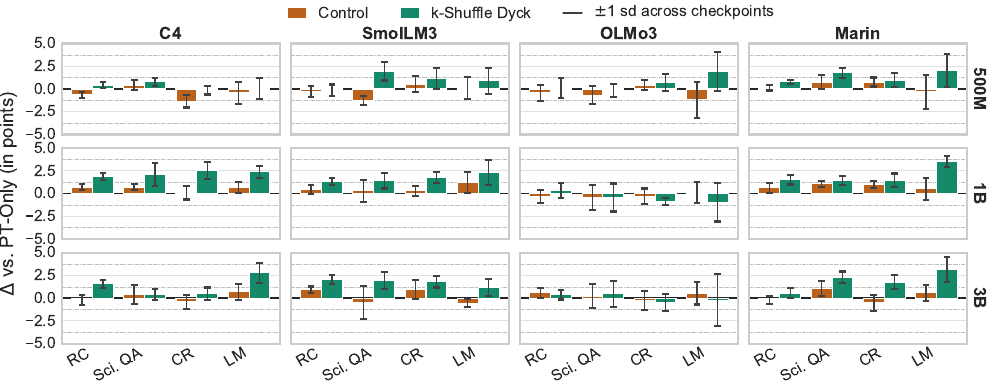}
    \caption{Downstream score changes from PPT relative to PT-Only.
    Bars show mean paired differences across 5K--10K checkpoints; whiskers denote $\pm$1 SD.
    Full scores are in Appendix Table~\ref{tab:appendix_section5_downstream_wide}.\looseness=-1}
    \label{fig:downstream}
\end{figure}

Figure~\ref{fig:downstream} shows the change in downstream performance from PPT relative to PT-Only.
$k$-Shuffle Dyck improves downstream performance in most scale-mixture pairs.
Of the 12 pairs (4 PT data mixtures $\times$ 3 model scales), 9 gain at least 0.6 points on the downstream average, with a mean gain of 1.6 among them.
The gain is also similar across scales (0.8 at 500M, 1.4 at 1B, 1.3 at 3B).
The effectiveness of PPT thus does not diminish as model capacity grows.
The remaining three pairs gain little or lose (C4 at 500M $+$0.3, OLMo3 at 1B $-$0.5, OLMo3 at 3B $+$0.0), which leaves OLMo3 as the only PT mixture that fails to benefit at more than one scale.
We examine its data composition in \S\ref{subsec:data_type}.\looseness=-1

The Control results indicate that the gains stem from the synthetic PPT data rather than from the additional optimization steps.
Control, which runs the same 500 warm-up steps on held-out PT text, stays close to PT-Only at every scale, with mean differences of $-$0.2 at 500M, $+$0.4 at 1B, and $+$0.2 at 3B.
In contrast, $k$-Shuffle Dyck outperforms Control in 10 of the 12 pairs, by mean margins of 1.0, 1.0, and 1.1 points at the three scales.

We observe that the downstream gains are also broad across task types.
Each category improves in 9 or more of the 12 scale-mixture pairs in Figure~\ref{fig:downstream} (RC 11, Science QA 10, CR 9, and LM 10).
In particular, language modeling gains the most at all three scales (1.2, 1.8, and 1.7 points, against 0.3 to 1.3 for the remaining categories).
At the benchmark level, ReCoRD and HellaSwag improve in all 12 pairs and LAMBADA in 10 (Appendix Table~\ref{tab:appendix_section5_downstream_wide}).
All three require information from the preceding passage, since their answers are often not determined by the candidate sentence alone.
We return to this shared requirement in \S\ref{subsec:linguistic}.

\begin{takeaway}
\textbf{Takeaway}\quad Synthetic PPT yields a downstream gain that holds across model scales and PT mixtures for a small fraction of the training cost.
\end{takeaway}

\subsection{Linguistic Competence}
\label{subsec:linguistic}
\begin{wraptable}{r}{0.4\textwidth}
\begin{center}
\small
\vspace{-4em}
\renewcommand{\arraystretch}{1.0}
\setlength{\tabcolsep}{3pt}
\caption{Verbatim retrieval as mean NLL ($\downarrow$) averaged over checkpoints at 5K--10K steps ($\Delta$ = PPT $-$ PT-Only).
Subscripts are standard deviations (SD) across checkpoints.
\colorbox{green!20}{Green} marks an improvement over the baseline.
}
\label{tab:section5_verbatim}
\resizebox{\linewidth}{!}{
\begin{tabular}{llcccc}
\toprule
 & & \textbf{PT-Only} & \multicolumn{2}{c}{\textbf{$\Delta$ NLL vs. PT-Only}} \\
\cmidrule(lr){4-5}
 & \textbf{Data Mix} & \textbf{NLL} & \textbf{Control} & \textbf{$k$-Shuffle Dyck} \\
\midrule
\multirow{4}{*}{\rotatebox{90}{500M}} & C4 & 3.358\textsubscript{.041} & +0.060\textsubscript{.044} & \cellcolor{green!20}{-0.064\textsubscript{.047}} \\
 & SmolLM3 & 3.391\textsubscript{.058} & +0.030\textsubscript{.033} & \cellcolor{green!20}{-0.020\textsubscript{.036}} \\
 & OLMo3 & 3.650\textsubscript{.063} & \cellcolor{green!20}{-0.046\textsubscript{.053}} & \cellcolor{green!20}{-0.066\textsubscript{.072}} \\
 & Marin & 3.341\textsubscript{.065} & \cellcolor{green!20}{-0.006\textsubscript{.016}} & \cellcolor{green!20}{-0.084\textsubscript{.033}} \\
\midrule
\multirow{4}{*}{\rotatebox{90}{1B}} & C4 & 3.307\textsubscript{.052} & \cellcolor{green!20}{-0.036\textsubscript{.031}} & \cellcolor{green!20}{-0.058\textsubscript{.027}} \\
 & SmolLM3 & 3.305\textsubscript{.073} & \cellcolor{green!20}{-0.033\textsubscript{.042}} & \cellcolor{green!20}{-0.058\textsubscript{.049}} \\
 & OLMo3 & 3.601\textsubscript{.093} & \cellcolor{green!20}{-0.053\textsubscript{.035}} & \cellcolor{green!20}{-0.023\textsubscript{.034}} \\
 & Marin & 3.233\textsubscript{.078} & \cellcolor{green!20}{-0.008\textsubscript{.030}} & \cellcolor{green!20}{-0.039\textsubscript{.021}} \\
\midrule
\multirow{4}{*}{\rotatebox{90}{3B}} & C4 & 3.227\textsubscript{.063} & +0.004\textsubscript{.019}
& \cellcolor{green!20}{-0.093\textsubscript{.028}} \\
 & SmolLM3 & 3.183\textsubscript{.075} & +0.031\textsubscript{.032} & \cellcolor{green!20}{-0.030\textsubscript{.046}} \\
 & OLMo3 & 3.673\textsubscript{.083} &  \cellcolor{green!20}{-0.115\textsubscript{.072}} & \cellcolor{green!20}{-0.056\textsubscript{.071}} \\
 & Marin & 3.150\textsubscript{.052} & \cellcolor{green!20}{-0.005\textsubscript{.026}} & \cellcolor{green!20}{-0.073\textsubscript{.024}} \\
\bottomrule
\end{tabular}
}
\end{center}
\vspace{-2em}
\end{wraptable}

\begin{figure}[t]
    \centering
    \includegraphics[width=0.99\linewidth]{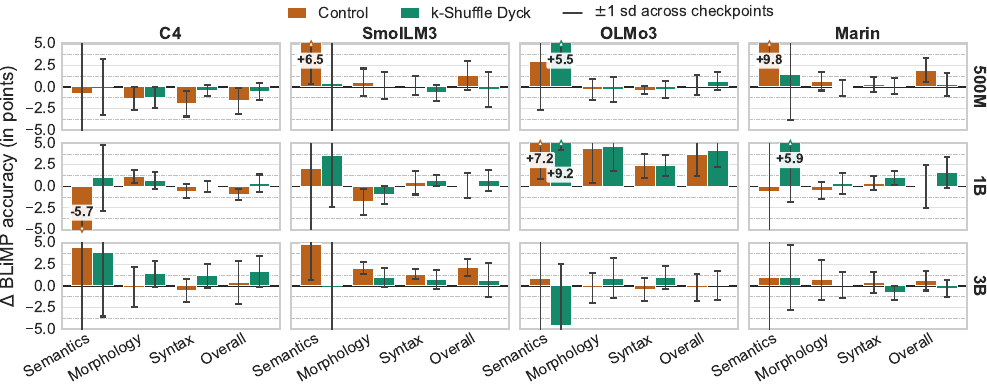}
    \caption{
    BLiMP score changes from PPT relative to PT-Only.
    Bars show mean paired differences across 5K--10K checkpoints; whiskers denote $\pm$1 SD.
    Full scores are in Appendix Table~\ref{tab:blimp_result_merged}.\looseness=-1
    }
    \label{fig:blimp}
\end{figure}

We now test whether the downstream gains from PPT stem from the grammatical prior proposed by \citet{hu-etal-2025-circuits}.
If they do, PPT should also improve grammatical acceptability on BLiMP.
Figure~\ref{fig:blimp} shows the BLiMP accuracy delta induced by PPT relative to PT-Only.
Although $k$-Shuffle Dyck improves overall accuracy in 9 of the 12 scale-mixture pairs, only one of these gains is stable (OLMo3 at 1B), against 10 of 12 for the downstream average (Appendix Table~\ref{tab:estimator_audit}).
This outcome does not align with prior work, which reports gains in grammatical acceptability from formal language PPT~\citep{hu-etal-2025-circuits,mita-etal-2026-language}.
Even at 500M, a scale within the range of prior work, we observe mixed results.
Two mixtures improve (OLMo3 $+$0.6, Marin $+$0.3), while two degrade (SmolLM3 $-$0.3, C4 $-$0.6).
The largest degradation is on C4, on which \citet{hu-etal-2025-circuits} report gains.
Looking at the results by model scale, we find that larger models do not resolve the disagreement either.
On Marin, the overall delta moves from $+$0.3 at 500M to $+$1.6 at 1B and back to $-$0.3 at 3B.

The subgroup breakdown does not change this picture.
If PPT imparts a grammatical prior, it should surface in Morphology and Syntax, which test hierarchical dependencies such as agreement and constituency.
These are the exact dependencies that $k$-Shuffle Dyck, a hierarchical bracket-matching language, is hypothesized to transfer.
Yet neither subgroup shows a stable gain (Appendix Table~\ref{tab:blimp_result_merged}).\looseness=-1

Verbatim retrieval, the second benchmark of \citet{hu-etal-2025-circuits}, shows a different pattern (Table~\ref{tab:section5_verbatim}).
$k$-Shuffle Dyck lowers NLL in all 12 scale-mixture pairs, with a stable reduction in 7 of them against 1 of 12 for BLiMP.
Control, by contrast, improves in 8 but only 3 of these reductions are stable.

We attribute this gap to the divergent demands of the two evaluations.
BLiMP compares two sentences that differ in one grammatical feature, where the deciding evidence lies within the sentence.
Verbatim retrieval instead requires locating a span earlier in a long context and copying it.
$k$-Shuffle Dyck requires an analogous operation over an abstract vocabulary, matching each closing bracket to its opening bracket across an arbitrary number of intervening symbols.
This suggests that PPT induces a long-range retrieval capability rather than the grammatical prior proposed by \citet{hu-etal-2025-circuits}.
It also explains why downstream gains are largest on LAMBADA, ReCoRD, and HellaSwag (\S\ref{subsec:general}), as all three require information from the preceding passage to determine an answer.
\begin{takeaway}
\textbf{Takeaway}\quad PPT improves long-range retrieval more reliably than grammatical acceptability, and its downstream gains, while broad, are largest on tasks that depend on the preceding context.\looseness=-1
\end{takeaway}

\section{Analysis}
\label{sec:analysis}

\paragraph{PT Data Type.}
\label{subsec:data_type}
\begin{table*}[t]
\small
\caption{
Ablations at 3B: (a) Marin composition and (b) PPT task.
Scores are mean paired differences from PT-Only over checkpoints at 5K--10K steps.
In (a), subscripts are standard deviations across checkpoints.
In (b), scores are averaged over the four mixtures, except for PT-Only, which gives absolute scores, and $n/4$ counts the mixtures in which the downstream average improves (per-mixture breakdown in Appendix Table~\ref{tab:appendix_ppt_tasks_per_mixture}).
\colorbox{green!20}{Green} marks an improvement.
}
\label{tab:section6_ablations}
\renewcommand{\arraystretch}{0.9}
\setlength{\tabcolsep}{3pt}
\begin{center}
\begin{tabular}{@{}c@{\hspace{0.2em}}lcccccccc}
\toprule
 & & \multicolumn{5}{c}{\textbf{Downstream (\%)}} & \textbf{BLiMP} & \textbf{Verbatim} & \\
\cmidrule(lr){3-7}\cmidrule(lr){8-8}\cmidrule(lr){9-9}
 & & RC & Sci. QA & CR & LM & \textbf{Avg} & (\%) & NLL $\downarrow$ & $n/4$ \\
\midrule
\multirow{4}{*}{(a)} & Marin (full) & \cellcolor{green!20}{+0.6\textsubscript{0.6}} & \cellcolor{green!20}{+2.3\textsubscript{0.6}} & \cellcolor{green!20}{+1.8\textsubscript{0.8}} & \cellcolor{green!20}{+3.1\textsubscript{1.3}} & \cellcolor{green!20}{+1.9\textsubscript{0.5}} & -0.3\textsubscript{1.0} & \cellcolor{green!20}{-0.073\textsubscript{0.024}} & -- \\
& 17\% Math & \cellcolor{green!20}{+1.0\textsubscript{0.4}} & \cellcolor{green!20}{+2.2\textsubscript{1.1}} & \cellcolor{green!20}{+1.8\textsubscript{0.4}} & \cellcolor{green!20}{+3.1\textsubscript{0.8}} & \cellcolor{green!20}{+2.0\textsubscript{0.3}} & -0.3\textsubscript{1.6} & \cellcolor{green!20}{-0.024\textsubscript{0.014}} & -- \\
& DCLM only & \cellcolor{green!20}{+0.8\textsubscript{0.2}} & \cellcolor{green!20}{+1.6\textsubscript{0.5}} & \cellcolor{green!20}{+1.6\textsubscript{0.9}} & \cellcolor{green!20}{+2.1\textsubscript{0.7}} & \cellcolor{green!20}{+1.5\textsubscript{0.3}} & +0.0\textsubscript{2.0} & \cellcolor{green!20}{-0.054\textsubscript{0.043}} & -- \\
& FineWeb-Edu only & \cellcolor{green!20}{+0.5\textsubscript{0.6}} & \cellcolor{green!20}{+0.0\textsubscript{1.3}} & \cellcolor{green!20}{+1.9\textsubscript{1.2}} & \cellcolor{green!20}{+1.5\textsubscript{1.4}} & \cellcolor{green!20}{+1.0\textsubscript{0.4}} & -1.5\textsubscript{3.9} & \cellcolor{green!20}{-0.042\textsubscript{0.050}} & -- \\
& Marin\textbackslash DCLM & \cellcolor{green!20}{+0.4\textsubscript{0.2}} & \cellcolor{green!20}{+0.6\textsubscript{0.9}} & -0.9\textsubscript{1.3} & \cellcolor{green!20}{+0.6\textsubscript{0.9}} & \cellcolor{green!20}{+0.2\textsubscript{0.5}} & \cellcolor{green!20}{+0.4\textsubscript{1.4}} & \cellcolor{green!20}{-0.012\textsubscript{0.041}} & -- \\
\midrule
\multirow{5}{*}{(b)} & PT-Only (absolute) & 50.7 & 50.1 & 55.9 & 40.8 & 49.4 & 79.0 & 3.308 & -- \\
& $k$-Shuffle Dyck & \cellcolor{green!20}{+1.1} & \cellcolor{green!20}{+1.3} & \cellcolor{green!20}{+0.9} & \cellcolor{green!20}{+1.7} & \cellcolor{green!20}{+1.3} & \cellcolor{green!20}{+0.5} & \cellcolor{green!20}{-0.063} & 3/4 \\
& MP-Struct Core & \cellcolor{green!20}{+0.7} & \cellcolor{green!20}{+0.9} & \cellcolor{green!20}{+0.7} & \cellcolor{green!20}{+1.5} & \cellcolor{green!20}{+0.9} & \cellcolor{green!20}{+0.6} & \cellcolor{green!20}{-0.044} & 3/4 \\
& NCA & \cellcolor{green!20}{+0.8} & \cellcolor{green!20}{+1.6} & \cellcolor{green!20}{+0.8} & \cellcolor{green!20}{+1.4} & \cellcolor{green!20}{+1.2} & \cellcolor{green!20}{+1.3} & \cellcolor{green!20}{-0.082} & 3/4 \\
& Set & -6.2 & -6.2 & -5.2 & -11.2 & -7.2 & -2.3 & +0.587 & 0/4 \\
\bottomrule
\end{tabular}
\end{center}
\end{table*}

Modern PT mixtures allocate a large share to code and mathematics, whose nested dependencies may already supply the structural signal that PPT provides (\S\ref{subsec:data_mixtures}).
If this holds, PPT should become redundant as code and mathematics content grows.
We test this on Marin, the mixture with the largest gains at 3B (\S\ref{subsec:general}; Figure~\ref{fig:downstream}), by varying its PT composition (Table~\ref{tab:section6_ablations}a).\footnote{\textbf{17\% Math} raises the math share from 1.3\% to 17.0\% at the expense of DCLM, leaving code unchanged at 6.1\%.
\textbf{DCLM only} keeps the web component alone, \textbf{FineWeb-Edu only} replaces it with the educational web corpus FineWeb-Edu, and \textbf{Marin\textbackslash DCLM} removes DCLM and renormalizes the remaining code and math.}
Raising the mathematics share from 1.3\% to 17.0\% lowers the web share from 92.6\% to 76.9\%, matching OLMo3 (Table~\ref{tab:data}).
This results in an average gain of 2.0 points against 1.9 for the Marin full PT mixture, with the per-category deltas almost unchanged.
Therefore, a 13$\times$ increase in mathematical content does not diminish the benefits of PPT.

By contrast, we find that web text is essential.
DCLM alone retains 1.5 of the 1.9 points on average and FineWeb-Edu retains 1.0.
Removing web text entirely (Marin\textbackslash DCLM in Table~\ref{tab:section6_ablations}a), which leaves 82\% source code and 18\% mathematics, reduces the average PPT gain to 0.2 points.
Language modeling follows the same pattern, gaining 1.5 to 3.1 points wherever web text is present and 0.6 without it.
Verbatim retrieval improves for all five PT data variants, and the improvement is stable in three of them (Marin full, 17\% Math, and DCLM only).
BLiMP echoes the pattern of \S\ref{subsec:linguistic}, in that the performance deltas remain small and none is stable.
The benefit of PPT thus holds across the diverse range of content that current PT mixtures occupy and disappears only under a web-free condition that no practical mixture approaches.

Neither the math share nor the web share explains the weak PPT gains on OLMo3 (\S\ref{subsec:general}).
The 17\% Math variant matches the 76.9\% web share of OLMo3 yet retains an average gain of 2.0 points.
Likewise, SmolLM3 differs from OLMo3 by only three points in web share, yet its PPT gain at 3B is 1.8 points against 0.0 for OLMo3 (Appendix Table~\ref{tab:appendix_ppt_tasks_per_mixture}; $k$-Shuffle Dyck).
We leave a direct ablation of OLMo3, and the property responsible for this gap, for future work.\footnote{The 3B experiments reported here already consume approximately 1.07T PT tokens, which places further runs beyond our computing capacity.}\looseness=-1
\begin{takeaway}
\textbf{Takeaway}\quad Code and mathematics do not make PPT redundant. Its gain holds across the compositions of current PT mixtures and requires only the presence of web text.
\end{takeaway}

\paragraph{PPT Tasks.}
\label{subsec:task_ablation}

To test whether the PPT benefit requires a formal grammar, we evaluate three further tasks at the 3B scale across all mixtures (Table~\ref{tab:section6_ablations}b): MP-Struct Core (formal), alongside NCA and Set (non-formal).
We first observe that the formal boundary does not separate the effective PPT tasks. MP-Struct Core ($+$0.9 on average) and the non-formal NCA ($+$1.2) yield gains comparable to $k$-Shuffle Dyck ($+$1.3), and all three improve 3 of the 4 mixtures.

In contrast, Set degrades downstream performance by 7.2 points and improves none. 
The three effective tasks all require retrieving a specific earlier position in the sequence.
Specifically, $k$-Shuffle Dyck matches a closing bracket to its opening bracket; MP-Struct Core anchors an explicit structural marker; and NCA reproduces the corresponding cell of the preceding automaton state.
Set only asks whether a token has appeared before.
The model can answer this by keeping a running record of seen tokens, without locating any specific earlier position.
This lack of positional retrieval may explain why Set performs poorly.\looseness=-1

This retrieval account is consistent with prior work, which already hints at dependency structure as the active property. Specifically, \citet{hu-etal-2025-circuits} require an effective PPT language to capture hierarchical dependencies, and \citet{mita-etal-2026-language} find that reducing retrieval ambiguity in those dependencies is what drives the gain.
Our results converge with prior work on which property of the task matters, and diverge on what the model acquires from it.
These results suggest that synthetic PPT aids general capability by inducing a long-range retrieval capability rather than a grammatical prior (\S\ref{subsec:linguistic}).\looseness=-1
\begin{takeaway}
\textbf{Takeaway}\quad Any PPT task that requires retrieving a specific earlier position is a safe choice, with $k$-Shuffle Dyck and NCA giving comparable gains. In contrast, Set, which only requires tracking seen tokens, degrades performance.
\end{takeaway}

\paragraph{Further Model and Data Scaling.}
\begin{wraptable}{r}{0.5\textwidth}
\small
\vspace{-2.25em}
\caption{Scaling. Each entry is the difference between PPT ($k$-Shuffle Dyck) and the PT-Only baseline at the same PT budget. Score breakdowns are in Appendix Tables~\ref{tab:appendix_scaling_absolute} and~\ref{tab:appendix_scaling_tasks}.
Results at 21B tokens are not comparable to \S\ref{sec:results}, as the two use different learning rate schedules.
}
\label{tab:section6_scaling}
\renewcommand{\arraystretch}{1.0}
\setlength{\tabcolsep}{1.5pt}
\setlength{\aboverulesep}{1.3pt}
\setlength{\belowrulesep}{1.3pt}
\begin{center}
\resizebox{\linewidth}{!}{
\begin{tabular}{lccccccccc}
\toprule
& \multicolumn{3}{c}{\textbf{C4 (3B)}} & \multicolumn{3}{c}{\textbf{Marin (3B)}} & \multicolumn{3}{c}{\textbf{Marin (7B)}} \\
\cmidrule(lr){2-4} \cmidrule(lr){5-7} \cmidrule(lr){8-10}
\textbf{Tokens} & \textbf{Avg} & \textbf{BLiMP} & \textbf{NLL $\downarrow$} & \textbf{Avg} & \textbf{BLiMP} & \textbf{NLL $\downarrow$} & \textbf{Avg} & \textbf{BLiMP} & \textbf{NLL $\downarrow$} \\
\midrule
21B & \cellcolor{green!20}+1.2 & \cellcolor{green!20}+3.0 & \cellcolor{green!20}-0.039 & \cellcolor{green!20}+1.1 & -3.1 & \cellcolor{green!20}-0.015 & \cellcolor{green!20}+0.5 & -0.2 & +0.008 \\
42B & \cellcolor{green!20}+0.7 & \cellcolor{green!20}+3.1 & +0.014 & \cellcolor{green!20}+1.6 & -10.3 & \cellcolor{green!20}-0.014 & \cellcolor{green!20}+0.3 & \cellcolor{green!20}+1.8 & +0.002 \\
63B & \cellcolor{green!20}+0.7 & -2.2 & \cellcolor{green!20}-0.033 & \cellcolor{green!20}+1.5 & -1.2 & \cellcolor{green!20}-0.019 & \cellcolor{green!20}+0.4 & -0.6 & \cellcolor{green!20}-0.004 \\
75.5B & -- & -- & -- & -- & -- & -- & \cellcolor{green!20}+1.0 & -1.1 & \cellcolor{green!20}-0.029 \\
84B & \cellcolor{green!20}+0.8 & -0.7 & \cellcolor{green!20}-0.021 & \cellcolor{green!20}+1.3 & \cellcolor{green!20}+1.3 & \cellcolor{green!20}-0.035 & -- & -- & -- \\
100B & \cellcolor{green!20}+1.6 & -15.5 & +0.008 & \cellcolor{green!20}+1.2 & -0.6 & \cellcolor{green!20}-0.039 & -- & -- & -- \\
\bottomrule
\end{tabular}
}
\end{center}
\vspace{-1em}
\end{wraptable}

Our main experiments stop at 21B PT tokens and at most 3B parameters.
The PPT benefit may still disappear with longer training (\S\ref{subsec:duration}), and larger models may not need it (\S\ref{subsec:scale}).
We therefore extend the 3B PT runs to 100B tokens on C4 and Marin, and train a 7B model on Marin for 75.5B tokens.
Table~\ref{tab:section6_scaling} shows that the downstream benefit persists under both extensions.
The downstream average improves in all 14 budget points across the three extended runs, by 1.0 on average on C4, 1.3 on Marin, and 0.6 at 7B.
Although PPT gains fluctuate, they do not degrade at longer budgets, ending at $+$1.6 on C4 and $+$1.2 on Marin at 100B, against $+$1.2 and $+$1.1 at 21B.
On Marin, PPT reaches 62.3 on average after 63B tokens, whereas PT-Only reaches 62.0 after 84B (Appendix Table~\ref{tab:appendix_scaling_absolute}). This saves at least 21B PT tokens.\looseness=-1

At 7B, the downstream gain narrows but persists.
On Marin, the average gain falls from 1.3 at 3B to 0.6 at 7B, yet it remains positive at every budget.
This narrowing may partly reflect budget rather than capacity.
The 7B runs reach roughly 11 tokens per parameter against 33 at 3B, and the largest 7B gain occurs at the longest budget ($+$1.0 at 75.5B tokens).
As in \S\ref{subsec:linguistic}, BLiMP gains remain mixed at both scales, improving in only 4 of the 14 budget points, whereas verbatim retrieval improves in 10.
Extended optimization and added capacity therefore preserve the downstream benefit of PPT.
\begin{takeaway}
\textbf{Takeaway}\quad PPT gains still persist up to 100B PT tokens and 7B parameters, and on Marin at 3B, PPT saves at least 21B PT tokens for about 1\% of the PT compute.
\end{takeaway}

\section{Conclusion}
We have presented the first systematic study of PPT at scale, spanning five PPT tasks, four PT data mixtures, four model scales (500M to 7B), and PT budgets of up to 100B tokens.
We first show that the downstream benefits of PPT persist under both parameter and PT budget scaling.
In contrast to previous work, we find that the attribution of these gains to a grammatical prior does not hold.
Specifically, downstream gains from PPT arise without equivalent gains in grammatical acceptability. We show instead that PPT is effective when the PPT task requires retrieving tokens from earlier positions in a sequence (\S\ref{sec:analysis}; PPT Tasks).
Finally, we demonstrate that PPT gains are insensitive to the share of code and math in the PT corpus, and diminish only when web text is absent.
These findings suggest that future PPT tasks may benefit more from targeting long-range retrieval directly than from mimicking the hierarchical structure of natural language grammar.

\section*{AI Use Statement}
In this work, we used Antigravity with Gemini 3.6/3.7/3.8 Flash to extend DataTrove preprocessing scripts for PPT data generation and to debug Nanotron scripts for deployment on our clusters (as they were originally designed for SmolLM model development).
Additionally, we used Gemini 3.1 Pro to assist with our literature survey.
Finally, we used Gemini 3.1 Pro, Claude Opus 5, and Gemini 3.8 Flash to improve the grammar and clarity of the draft.

\section*{Reproducibility Statement}
Our code and a step-by-step guide for preprocessing, training, and evaluation for all approaches are available in our GitHub repository: \githubdown~\url{https://github.com/gucci-j/verify-ppt-at-scale}.
Full details on hyperparameters, software, and hardware, including specific versions used, are provided in Appendix \ref{app:setup}.
All artifacts, including pre-processed datasets and model checkpoints, are available on the Hugging Face Hub: \huggingfacedown~\url{https://huggingface.co/verify-ppt}.

\section*{Acknowledgments}
We acknowledge
(1) IT Services at the University of Sheffield for the provision of high-performance computing services; 
(2) the Isambard-AI National AI Research Resource (AIRR), operated by the University of Bristol and funded by the UK Government’s Department for Science, Innovation and Technology (DSIT) via UK Research and Innovation, under Science and Technology Facilities Council grant [ST/AIRR/I-A-I/1023]; 
(3) the EuroHPC Joint Undertaking for awarding access to Leonardo (hosted by CINECA, Italy) and JUPITER (hosted by JSC, Germany); and 
(4) the use of ``mdx: a platform for building data-empowered society''~\citep{9927975}.
This work was supported by the ``Development Acceleration Use'' program of ABCI 3.0, provided by AIST and AIST Solutions and by the ``R\&D Hub Aimed at Ensuring Transparency and Reliability of Generative AI Models'' project of the Ministry of Education, Culture, Sports, Science and Technology of Japan.
AY is supported by the Engineering and Physical Sciences Research Council (EPSRC) [grant number EP/W524360/1] and the Japan Student Services Organization (JASSO) Student Exchange Support Program (Graduate Scholarship for Degree Seeking Students).

\bibliography{anthology-1,anthology-2,custom}
\bibliographystyle{iclr2027_conference}
\clearpage

\begin{tcolorbox}[
    title=Appendix Directory,
    colback=lightgray!10,
    colframe=black,
    fonttitle=\bfseries, 
    rounded corners
]
\begin{itemize}[leftmargin=*]
    \item \textbf{Appendix A:} \hyperref[app:setup]{Supplementary Experimental Setup}
        \begin{itemize}
            \item \hyperref[app:preprocessing_details]{Pre-processing Details}
            \item \hyperref[app:training_details]{Training Details}
            \item \hyperref[app:evaluation_details]{Evaluation Details}
        \end{itemize}

    \item \textbf{Appendix B:} \hyperref[app:suppl_results]{Supplementary Results}
        \begin{itemize}
            \item \hyperref[app:estimator]{Checkpoint Sensitivity of Single-Snapshot Estimates}
            \item \hyperref[app:seed]{Sensitivity to Random Seeds}
        \end{itemize}

    \item \textbf{Appendix C:} \hyperref[app:suppl_analysis]{Supplementary Analysis}
\end{itemize}
\end{tcolorbox}

\begin{table}[ht]
\small
\caption{Generation parameters for the pre-pretraining corpora.}
\label{tab:ppt_generation}
\begin{center}
\begin{tabular}{llr}
\toprule
\textbf{Task} & \textbf{Symbol inventory} & \textbf{Ints / doc}\\
\midrule
$k$-Shuffle Dyck & $k = 64$, $p_\text{open} = 0.50$, depth $\in [1,8]$ & 2,048\\
MP-Struct Core  & $k_\text{struct} = 1$, $k_\text{dep} = 4$, 3 heads, 1 leaf & 2,048\\
Set             & 999 values, separator 999 & 2,048\\
NCA             & $12{\times}12$ grid, 10 states, $2{\times}2$ patches, $T = 10^{-4}$ & 1,406 \\
\bottomrule
\end{tabular}
\end{center}
\end{table}

\appendix
\section{Supplementary Experimental Setup}
\label{app:setup}

\subsection{Pre-processing Details}
\label{app:preprocessing_details}

Each synthetic corpus is emitted as space-separated integers and tokenized with the SmolLM3 tokenizer (\S\ref{subsec:training_pipeline}).
Symbol identifiers remain constrained to three digits. 
This design ensures that each integer maps to a fixed number of BPE pieces, enabling each document to occupy exactly one 4,096-token sequence.
Table~\ref{tab:ppt_generation} lists the generation parameters.
Note that Control uses the same tokenization and packing pipeline as $\mathcal{D}_\text{PT}$.

\paragraph{$k$-Shuffle Dyck.}
We use the generator of \citet{hu-etal-2025-circuits} without modification, constraining bracket depth to $[1, 8]$.

\paragraph{MP-Struct Core.}
We use the generator released by \citet{mita-etal-2026-language} unchanged,\footnote{\url{https://github.com/osekilab/LAD-PPT}} retaining the default configuration of one structural bracket type, four dependency types, active functional head markers, and disabled complex arguments.
We increase document length from the default of 1,024 to 2,048 integers, allowing each document to populate the 4,096-token context window.

\paragraph{Set.}
Following \citet{jiang2026procedural}, each document concatenates an input sequence, a separator, and the unique input values in order of initial appearance.

\paragraph{NCA.}
We implement the configuration identified as optimal by \citet{lee2026traininglanguagemodelsneural}, retaining only trajectories with a GZIP compression ratio of at least 0.50.
Because patch identifiers reach five digits, subword tokenization introduces slight length variation: documents average 4,085.6 tokens (SD 206.7) rather than matching the context boundary uniformly.

\subsection{Training Details}
\label{app:training_details}

Table~\ref{tab:model_hyperparameters} summarizes architectural specifications and training hyperparameters across the four parameter scales evaluated in this study (500M, 1B, 3B, and 7B).

\begin{table*}[t]
\begin{center}
\small
\caption{Architectural and optimization hyperparameters across model scales.}
\label{tab:model_hyperparameters}
\resizebox{\linewidth}{!}{
\begin{tabular}{lcccc}
\toprule
\textbf{Hyperparameter} & \textbf{500M} & \textbf{1B} & \textbf{3B} & \textbf{7B} \\
\midrule
\multicolumn{5}{l}{\textit{Architecture}} \\
Hidden size ($d_{\text{model}}$) & 1,024 & 1,536 & 2,560 & 3,584 \\
Intermediate size ($d_{\text{ff}}$) & 2,816 & 4,096 & 6,912 & 18,944 \\
Number of layers ($n_{\text{layer}}$) & 32 & 36 & 40 & 28 \\
Attention heads ($n_{\text{head}}$) & 16 & 12 & 20 & 28 \\
Key-Value heads ($n_{\text{kv}}$) & 4 & 4 & 4 & 4 \\
Vocabulary size & 128,256 & 128,256 & 128,256 & 128,256 \\
Max context length & 4,096 & 4,096 & 4,096 & 4,096 \\
RoPE base frequency ($\theta$) & 50,000 & 50,000 & 50,000 & 50,000 \\
RMSNorm $\epsilon$ & $1.0 \times 10^{-5}$ & $1.0 \times 10^{-5}$ & $1.0 \times 10^{-5}$ & $1.0 \times 10^{-6}$ \\
Weight tying & True & True & True & True \\
\midrule
\multicolumn{5}{l}{\textit{Optimization \& LR Schedule (WSD)}} \\
Optimizer & AdamW & AdamW & AdamW & AdamW \\
$\beta_1, \beta_2$ & 0.9, 0.95 & 0.9, 0.95 & 0.9, 0.95 & 0.9, 0.95 \\
$\epsilon_{\text{adam}}$ & $1.0 \times 10^{-8}$ & $1.0 \times 10^{-8}$ & $1.0 \times 10^{-8}$ & $1.0 \times 10^{-8}$ \\
Weight decay & 0.1 & 0.1 & 0.1 & 0.1 \\
Gradient clipping & 1.0 & 1.0 & 1.0 & 1.0 \\
LR schedule & WSD & WSD & WSD & WSD \\
LR warmup style & Linear & Linear & Linear & Linear \\
LR warmup steps (21B / 100B or 75.5B) & 1,000 / --- & 1,000 / --- & 1,000 / 2,000 & 1,000 / 2,000 \\
LR decay style & Cosine & Cosine & Cosine & Cosine \\
LR decay duration & Final 10\% & Final 10\% & Final 10\% & Final 10\% \\
Peak learning rate ($\eta_{\max}$) & $5.0 \times 10^{-4}$ & $5.0 \times 10^{-4}$ & $5.0 \times 10^{-4}$ & $3.0 \times 10^{-4}$ \\
Min learning rate ($\eta_{\min}$) & $5.0 \times 10^{-5}$ & $5.0 \times 10^{-5}$ & $5.0 \times 10^{-5}$ & $3.0 \times 10^{-5}$ \\
Global batch size (sequences) & 512 & 512 & 512 & 512 \\
Global batch size (tokens) & $\approx 2.1\text{M}$ & $\approx 2.1\text{M}$ & $\approx 2.1\text{M}$ & $\approx 2.1\text{M}$ \\
Precision & \texttt{bfloat16} & \texttt{bfloat16} & \texttt{bfloat16} & \texttt{bfloat16} \\
\midrule
\multicolumn{5}{l}{\textit{Training Budgets}} \\
PPT duration (steps / tokens) & 500 / $\approx 1.05\text{B}$ & 500 / $\approx 1.05\text{B}$ & 500 / $\approx 1.05\text{B}$ & 500 / $\approx 1.05\text{B}$ \\
PT standard (steps / tokens) & 10,000 / $\approx 21\text{B}$ & 10,000 / $\approx 21\text{B}$ & 10,000 / $\approx 21\text{B}$ & 10,000 / $\approx 21\text{B}$ \\
PT extended (steps / tokens) & --- & --- & 47,684 / $\approx 100\text{B}$ & 36,000 / $\approx 75.5\text{B}$ \\
\bottomrule
\end{tabular}
}
\end{center}
\end{table*}

We run training with Nanotron (v0.4)\footnote{\url{https://github.com/huggingface/nanotron}} and FlashAttention-2~\citep[v2.7.4.post1]{dao2024flashattention}.
Data processing relies on DataTrove~\citep[v0.4.0]{Penedo_DataTrove_large_scale} and \texttt{datasets}~\citep[v5.0.1]{lhoest-etal-2021-datasets}.
To minimize GPU memory overhead during large-batch optimization, we employ \texttt{flash\_adamw} from the FlashOptim library~\citep[v0.1.4]{ortiz2026flashoptim}.
Due to resource constraints, we run experiments on multiple NVIDIA GPU clusters, equipped with A100, H100, H200, or GH200 devices.
To ensure cross-platform reproducibility, all runs are executed within a uniform containerized environment built upon the NVIDIA vLLM container (\texttt{nvcr.io/nvidia/vllm:26.01-py3})\footnote{\url{https://catalog.ngc.nvidia.com/orgs/nvidia/-/containers/vllm/26.01-py3}} with PyTorch 2.10.0 and CUDA 13.1.

\paragraph{Compute Cost.}
On 16 A100 40GB GPUs, the 500-step PPT requires 3 hours and 32 minutes (approximately 56.5 GPU hours) for the 3B parameter model.
The corresponding 21B-token pre-training phase consumes 2 days and 23 hours (approximately 1,136 GPU hours).
Consequently, the PPT stage represents 4.74\% of total wall-clock execution time, closely matching the theoretical projection of 5.0\%.

\subsection{Evaluation Details}
\label{app:evaluation_details}
All evaluations except verbatim retrieval use \texttt{lm-evaluation-harness}~\citep[v0.4.10]{eval-harness} with default settings, automated batch sizing, and \texttt{bfloat16} precision.

\paragraph{BLiMP.}
We adopt the twelve paradigm groups and their assignment to Semantics, Morphology, and Syntax from \citet{warstadt-etal-2020-blimp-benchmark}.

\paragraph{Verbatim Retrieval.}
We evaluate verbatim retrieval using the five repeat-condition stimulus sets of \citet{armeni-etal-2022-characterizing}, released as \texttt{categorized\_lists\_sce1--5\_repeat},\footnote{\url{https://github.com/KristijanArmeni/verbatim-memory-in-NLMs/tree/main/data/rnn_input_files}} wherein each context contains two presentations of a noun list.
Word-level markers distinguish the first occurrence from the repeat.
We score each stimulus in a single teacher-forced forward pass and map word markers onto BPE tokens by character offset.
Following \citet{hu-etal-2025-circuits}, we report the mean NLL over the full stimulus rather than over the repeated span alone, averaged over stimuli.

\section{Supplementary Results}
\label{app:suppl_results}

\begin{table*}[!t]
\tiny
\caption{Task-level downstream performance. \colorbox{gray!20}{Gray} rows give the PT-Only mean over checkpoints at 5K--10K steps; the rows below give the mean paired difference from that baseline, with standard deviations across checkpoints as subscripts. \colorbox{green!20}{Green} marks an improvement.}
\label{tab:appendix_section5_downstream_wide}
\begin{center}
\setlength{\aboverulesep}{1.3pt}
\setlength{\belowrulesep}{1.3pt}
\renewcommand{\arraystretch}{1.0}
\setlength{\tabcolsep}{3pt}
\resizebox{\textwidth}{!}{
\begin{tabular}{llcccccccccc}
\toprule
 &  & \multicolumn{2}{c}{\textbf{RC}} & \multicolumn{3}{c}{\textbf{Science QA}} & \multicolumn{4}{c}{\textbf{CR}} & \multicolumn{1}{c}{\textbf{LM}} \\
\cmidrule(lr){3-4} \cmidrule(lr){5-7} \cmidrule(lr){8-11} \cmidrule(lr){12-12}
 & \textbf{Approach} & \textbf{RACE} & \textbf{ReCoRD} & \textbf{SciQ} & \textbf{ARC-Easy} & \textbf{OpenBookQA} & \textbf{COPA} & \textbf{PIQA} & \textbf{SIQA} & \textbf{HellaSwag} & \textbf{LAMBADA} \\
\midrule
\multirow{12}{*}{\rotatebox[origin=c]{90}{\textbf{500M}}} & \cellcolor{gray!20}C4 (PT-Only) & \cellcolor{gray!20}29.3 & \cellcolor{gray!20}66.8 & \cellcolor{gray!20}61.5 & \cellcolor{gray!20}35.9 & \cellcolor{gray!20}27.4 & \cellcolor{gray!20}71.2 & \cellcolor{gray!20}67.8 & \cellcolor{gray!20}35.1 & \cellcolor{gray!20}37.7 & \cellcolor{gray!20}30.8 \\
 & Control & -0.5~\textsubscript{0.6} & -0.8~\textsubscript{0.5} & -0.4~\textsubscript{1.6} & \cellcolor{green!20}{+1.4~\textsubscript{1.2}} & \cellcolor{green!20}{+0.2~\textsubscript{1.1}} & -4.5~\textsubscript{2.8} & -1.0~\textsubscript{0.6} & \cellcolor{green!20}{+0.6~\textsubscript{0.7}} & -0.6~\textsubscript{0.6} & -0.4~\textsubscript{1.2} \\
 & $k$-Shuffle Dyck & -0.0~\textsubscript{1.0} & \cellcolor{green!20}{+0.9~\textsubscript{0.7}} & \cellcolor{green!20}{+1.1~\textsubscript{1.3}} & \cellcolor{green!20}{+0.6~\textsubscript{1.1}} & \cellcolor{green!20}{+0.7~\textsubscript{0.9}} & -2.3~\textsubscript{2.1} & -1.0~\textsubscript{1.1} & \cellcolor{green!20}{+1.0~\textsubscript{0.6}} & \cellcolor{green!20}{+1.7~\textsubscript{0.7}} & \cellcolor{green!20}{+0.1~\textsubscript{1.1}} \\
\cmidrule(lr){2-12}
 & \cellcolor{gray!20}SmolLM3 (PT-Only) & \cellcolor{gray!20}29.6 & \cellcolor{gray!20}65.1 & \cellcolor{gray!20}67.8 & \cellcolor{gray!20}53.5 & \cellcolor{gray!20}30.0 & \cellcolor{gray!20}66.0 & \cellcolor{gray!20}64.9 & \cellcolor{gray!20}38.3 & \cellcolor{gray!20}36.3 & \cellcolor{gray!20}35.7 \\
 & Control & -0.2~\textsubscript{0.8} & -0.3~\textsubscript{0.7} & -1.6~\textsubscript{1.5} & +0.0~\textsubscript{0.9} & -2.2~\textsubscript{1.5} & \cellcolor{green!20}{+2.8~\textsubscript{2.4}} & -0.5~\textsubscript{1.3} & \cellcolor{green!20}{+0.6~\textsubscript{0.5}} & -0.8~\textsubscript{0.4} & \cellcolor{green!20}{+0.1~\textsubscript{1.2}} \\
 & $k$-Shuffle Dyck & -0.8~\textsubscript{1.0} & \cellcolor{green!20}{+0.5~\textsubscript{0.9}} & \cellcolor{green!20}{+2.8~\textsubscript{1.7}} & \cellcolor{green!20}{+2.6~\textsubscript{2.0}} & \cellcolor{green!20}{+0.4~\textsubscript{0.7}} & \cellcolor{green!20}{+3.0~\textsubscript{4.5}} & \cellcolor{green!20}{+0.3~\textsubscript{0.8}} & \cellcolor{green!20}{+0.7~\textsubscript{0.6}} & \cellcolor{green!20}{+0.8~\textsubscript{0.4}} & \cellcolor{green!20}{+0.9~\textsubscript{1.4}} \\
\cmidrule(lr){2-12}
 & \cellcolor{gray!20}OLMo3 (PT-Only) & \cellcolor{gray!20}28.2 & \cellcolor{gray!20}60.7 & \cellcolor{gray!20}58.5 & \cellcolor{gray!20}38.1 & \cellcolor{gray!20}26.5 & \cellcolor{gray!20}67.4 & \cellcolor{gray!20}62.7 & \cellcolor{gray!20}38.3 & \cellcolor{gray!20}33.4 & \cellcolor{gray!20}34.2 \\
 & Control & -0.1~\textsubscript{1.3} & -0.7~\textsubscript{0.6} & -2.9~\textsubscript{2.7} & \cellcolor{green!20}{+0.2~\textsubscript{1.5}} & \cellcolor{green!20}{+0.6~\textsubscript{1.4}} & \cellcolor{green!20}{+1.4~\textsubscript{1.7}} & \cellcolor{green!20}{+0.2~\textsubscript{1.2}} & +0.0~\textsubscript{1.0} & \cellcolor{green!20}{+0.1~\textsubscript{0.3}} & -1.2~\textsubscript{2.0} \\
 & $k$-Shuffle Dyck & -0.1~\textsubscript{1.8} & \cellcolor{green!20}{+0.4~\textsubscript{0.6}} & -0.0~\textsubscript{1.5} & \cellcolor{green!20}{+0.1~\textsubscript{0.8}} & -0.6~\textsubscript{1.0} & \cellcolor{green!20}{+1.6~\textsubscript{2.5}} & \cellcolor{green!20}{+0.3~\textsubscript{1.0}} & \cellcolor{green!20}{+0.2~\textsubscript{0.8}} & \cellcolor{green!20}{+0.8~\textsubscript{0.2}} & \cellcolor{green!20}{+1.9~\textsubscript{2.2}} \\
\cmidrule(lr){2-12}
 & \cellcolor{gray!20}Marin (PT-Only) & \cellcolor{gray!20}29.6 & \cellcolor{gray!20}67.6 & \cellcolor{gray!20}66.8 & \cellcolor{gray!20}51.1 & \cellcolor{gray!20}28.7 & \cellcolor{gray!20}69.0 & \cellcolor{gray!20}66.2 & \cellcolor{gray!20}38.6 & \cellcolor{gray!20}36.5 & \cellcolor{gray!20}38.7 \\
 & Control & \cellcolor{green!20}{+0.4~\textsubscript{0.4}} & -0.1~\textsubscript{0.5} & \cellcolor{green!20}{+1.6~\textsubscript{1.0}} & \cellcolor{green!20}{+0.1~\textsubscript{1.1}} & \cellcolor{green!20}{+0.6~\textsubscript{0.9}} & \cellcolor{green!20}{+1.7~\textsubscript{1.9}} & \cellcolor{green!20}{+0.6~\textsubscript{0.7}} & \cellcolor{green!20}{+0.5~\textsubscript{0.7}} & \cellcolor{green!20}{+0.3~\textsubscript{0.2}} & -0.3~\textsubscript{1.9} \\
 & $k$-Shuffle Dyck & \cellcolor{green!20}{+0.6~\textsubscript{0.4}} & \cellcolor{green!20}{+0.9~\textsubscript{0.4}} & \cellcolor{green!20}{+1.9~\textsubscript{1.7}} & \cellcolor{green!20}{+2.5~\textsubscript{0.3}} & \cellcolor{green!20}{+0.9~\textsubscript{0.8}} & \cellcolor{green!20}{+1.0~\textsubscript{3.0}} & \cellcolor{green!20}{+0.3~\textsubscript{0.7}} & \cellcolor{green!20}{+0.8~\textsubscript{0.4}} & \cellcolor{green!20}{+1.8~\textsubscript{0.3}} & \cellcolor{green!20}{+2.0~\textsubscript{1.8}} \\
\midrule
\multirow{12}{*}{\rotatebox[origin=c]{90}{\textbf{1B}}} & \cellcolor{gray!20}C4 (PT-Only) & \cellcolor{gray!20}30.1 & \cellcolor{gray!20}69.9 & \cellcolor{gray!20}64.3 & \cellcolor{gray!20}37.6 & \cellcolor{gray!20}28.6 & \cellcolor{gray!20}68.5 & \cellcolor{gray!20}68.8 & \cellcolor{gray!20}35.8 & \cellcolor{gray!20}42.2 & \cellcolor{gray!20}33.6 \\
 & Control & \cellcolor{green!20}{+0.7~\textsubscript{0.5}} & \cellcolor{green!20}{+0.7~\textsubscript{0.3}} & \cellcolor{green!20}{+0.6~\textsubscript{0.9}} & \cellcolor{green!20}{+1.8~\textsubscript{0.8}} & -0.2~\textsubscript{0.8} & -0.5~\textsubscript{2.6} & \cellcolor{green!20}{+0.1~\textsubscript{0.8}} & \cellcolor{green!20}{+0.6~\textsubscript{0.6}} & \cellcolor{green!20}{+0.3~\textsubscript{0.3}} & \cellcolor{green!20}{+0.6~\textsubscript{0.6}} \\
 & $k$-Shuffle Dyck & \cellcolor{green!20}{+1.1~\textsubscript{0.4}} & \cellcolor{green!20}{+2.7~\textsubscript{0.4}} & \cellcolor{green!20}{+2.2~\textsubscript{1.1}} & \cellcolor{green!20}{+3.4~\textsubscript{2.6}} & \cellcolor{green!20}{+0.7~\textsubscript{1.0}} & \cellcolor{green!20}{+4.3~\textsubscript{2.7}} & \cellcolor{green!20}{+1.1~\textsubscript{0.8}} & \cellcolor{green!20}{+1.4~\textsubscript{0.9}} & \cellcolor{green!20}{+3.4~\textsubscript{0.5}} & \cellcolor{green!20}{+2.4~\textsubscript{0.7}} \\
\cmidrule(lr){2-12}
 & \cellcolor{gray!20}SmolLM3 (PT-Only) & \cellcolor{gray!20}30.9 & \cellcolor{gray!20}68.3 & \cellcolor{gray!20}70.1 & \cellcolor{gray!20}57.4 & \cellcolor{gray!20}32.0 & \cellcolor{gray!20}67.2 & \cellcolor{gray!20}66.8 & \cellcolor{gray!20}39.9 & \cellcolor{gray!20}39.5 & \cellcolor{gray!20}38.5 \\
 & Control & \cellcolor{green!20}{+0.1~\textsubscript{1.2}} & \cellcolor{green!20}{+0.8~\textsubscript{0.4}} & \cellcolor{green!20}{+2.1~\textsubscript{1.4}} & \cellcolor{green!20}{+0.6~\textsubscript{0.6}} & -1.9~\textsubscript{2.2} & \cellcolor{green!20}{+0.5~\textsubscript{3.0}} & \cellcolor{green!20}{+0.3~\textsubscript{1.4}} & -0.5~\textsubscript{0.7} & \cellcolor{green!20}{+1.0~\textsubscript{0.3}} & \cellcolor{green!20}{+1.2~\textsubscript{1.2}} \\
 & $k$-Shuffle Dyck & \cellcolor{green!20}{+0.3~\textsubscript{0.8}} & \cellcolor{green!20}{+2.4~\textsubscript{0.2}} & \cellcolor{green!20}{+1.4~\textsubscript{1.7}} & \cellcolor{green!20}{+2.7~\textsubscript{1.0}} & \cellcolor{green!20}{+0.2~\textsubscript{1.2}} & \cellcolor{green!20}{+3.0~\textsubscript{3.6}} & \cellcolor{green!20}{+1.3~\textsubscript{0.7}} & -0.1~\textsubscript{0.6} & \cellcolor{green!20}{+2.9~\textsubscript{0.4}} & \cellcolor{green!20}{+2.3~\textsubscript{1.3}} \\
\cmidrule(lr){2-12}
 & \cellcolor{gray!20}OLMo3 (PT-Only) & \cellcolor{gray!20}28.7 & \cellcolor{gray!20}61.8 & \cellcolor{gray!20}59.9 & \cellcolor{gray!20}39.8 & \cellcolor{gray!20}26.8 & \cellcolor{gray!20}68.3 & \cellcolor{gray!20}64.0 & \cellcolor{gray!20}39.1 & \cellcolor{gray!20}35.0 & \cellcolor{gray!20}35.7 \\
 & Control & -0.9~\textsubscript{1.2} & \cellcolor{green!20}{+0.3~\textsubscript{0.8}} & -0.8~\textsubscript{1.2} & -0.6~\textsubscript{1.1} & +0.0~\textsubscript{2.7} & -0.2~\textsubscript{3.2} & -0.6~\textsubscript{0.7} & -0.5~\textsubscript{0.8} & \cellcolor{green!20}{+0.1~\textsubscript{0.4}} & \cellcolor{green!20}{+0.1~\textsubscript{1.2}} \\
 & $k$-Shuffle Dyck & \cellcolor{green!20}{+0.4~\textsubscript{1.4}} & \cellcolor{green!20}{+0.3~\textsubscript{0.7}} & -1.0~\textsubscript{2.3} & -0.4~\textsubscript{1.3} & \cellcolor{green!20}{+0.1~\textsubscript{2.9}} & -2.8~\textsubscript{1.5} & -0.0~\textsubscript{0.4} & -1.2~\textsubscript{0.7} & \cellcolor{green!20}{+0.5~\textsubscript{0.7}} & -0.9~\textsubscript{2.1} \\
\cmidrule(lr){2-12}
 & \cellcolor{gray!20}Marin (PT-Only) & \cellcolor{gray!20}30.9 & \cellcolor{gray!20}71.1 & \cellcolor{gray!20}70.2 & \cellcolor{gray!20}56.0 & \cellcolor{gray!20}31.9 & \cellcolor{gray!20}69.8 & \cellcolor{gray!20}68.1 & \cellcolor{gray!20}41.0 & \cellcolor{gray!20}41.6 & \cellcolor{gray!20}44.0 \\
 & Control & -0.1~\textsubscript{0.9} & \cellcolor{green!20}{+1.3~\textsubscript{0.5}} & \cellcolor{green!20}{+3.0~\textsubscript{0.9}} & \cellcolor{green!20}{+0.9~\textsubscript{1.0}} & -0.7~\textsubscript{1.0} & \cellcolor{green!20}{+2.7~\textsubscript{2.0}} & \cellcolor{green!20}{+0.5~\textsubscript{0.9}} & -0.1~\textsubscript{0.4} & \cellcolor{green!20}{+0.9~\textsubscript{0.3}} & \cellcolor{green!20}{+0.5~\textsubscript{1.2}} \\
 & $k$-Shuffle Dyck & \cellcolor{green!20}{+0.5~\textsubscript{1.2}} & \cellcolor{green!20}{+2.5~\textsubscript{0.3}} & \cellcolor{green!20}{+1.7~\textsubscript{1.5}} & \cellcolor{green!20}{+2.6~\textsubscript{0.7}} & \cellcolor{green!20}{+0.1~\textsubscript{0.5}} & \cellcolor{green!20}{+2.0~\textsubscript{2.4}} & \cellcolor{green!20}{+0.5~\textsubscript{0.7}} & \cellcolor{green!20}{+0.6~\textsubscript{0.3}} & \cellcolor{green!20}{+2.8~\textsubscript{0.3}} & \cellcolor{green!20}{+3.5~\textsubscript{0.6}} \\
\midrule
\multirow{12}{*}{\rotatebox[origin=c]{90}{\textbf{3B}}} & \cellcolor{gray!20}C4 (PT-Only) & \cellcolor{gray!20}31.8 & \cellcolor{gray!20}74.4 & \cellcolor{gray!20}66.6 & \cellcolor{gray!20}41.4 & \cellcolor{gray!20}29.7 & \cellcolor{gray!20}72.8 & \cellcolor{gray!20}71.3 & \cellcolor{gray!20}36.5 & \cellcolor{gray!20}48.0 & \cellcolor{gray!20}36.7 \\
 & Control & -0.1~\textsubscript{0.8} & -0.3~\textsubscript{0.7} & \cellcolor{green!20}{+0.8~\textsubscript{1.3}} & \cellcolor{green!20}{+0.8~\textsubscript{2.2}} & -0.3~\textsubscript{1.7} & -1.0~\textsubscript{1.8} & -0.2~\textsubscript{0.9} & -0.4~\textsubscript{0.9} & \cellcolor{green!20}{+0.1~\textsubscript{0.3}} & \cellcolor{green!20}{+0.7~\textsubscript{0.9}} \\
 & $k$-Shuffle Dyck & \cellcolor{green!20}{+1.3~\textsubscript{1.3}} & \cellcolor{green!20}{+1.8~\textsubscript{0.6}} & \cellcolor{green!20}{+2.1~\textsubscript{1.1}} & -1.3~\textsubscript{2.3} & \cellcolor{green!20}{+0.5~\textsubscript{0.8}} & -1.3~\textsubscript{2.5} & \cellcolor{green!20}{+0.3~\textsubscript{0.5}} & -0.2~\textsubscript{0.8} & \cellcolor{green!20}{+3.2~\textsubscript{0.4}} & \cellcolor{green!20}{+2.8~\textsubscript{1.1}} \\
\cmidrule(lr){2-12}
 & \cellcolor{gray!20}SmolLM3 (PT-Only) & \cellcolor{gray!20}31.1 & \cellcolor{gray!20}71.9 & \cellcolor{gray!20}74.9 & \cellcolor{gray!20}61.9 & \cellcolor{gray!20}33.7 & \cellcolor{gray!20}68.8 & \cellcolor{gray!20}68.7 & \cellcolor{gray!20}41.6 & \cellcolor{gray!20}44.9 & \cellcolor{gray!20}44.5 \\
 & Control & \cellcolor{green!20}{+1.2~\textsubscript{0.7}} & \cellcolor{green!20}{+0.6~\textsubscript{0.2}} & -0.1~\textsubscript{1.2} & \cellcolor{green!20}{+0.1~\textsubscript{3.9}} & -1.4~\textsubscript{1.7} & \cellcolor{green!20}{+3.0~\textsubscript{3.7}} & \cellcolor{green!20}{+0.5~\textsubscript{0.7}} & -0.4~\textsubscript{1.1} & \cellcolor{green!20}{+0.7~\textsubscript{0.3}} & -0.5~\textsubscript{0.4} \\
 & $k$-Shuffle Dyck & \cellcolor{green!20}{+2.5~\textsubscript{1.0}} & \cellcolor{green!20}{+1.6~\textsubscript{0.4}} & \cellcolor{green!20}{+2.8~\textsubscript{1.7}} & \cellcolor{green!20}{+2.3~\textsubscript{1.1}} & \cellcolor{green!20}{+0.8~\textsubscript{1.8}} & \cellcolor{green!20}{+2.3~\textsubscript{2.4}} & \cellcolor{green!20}{+1.2~\textsubscript{0.6}} & \cellcolor{green!20}{+0.3~\textsubscript{0.4}} & \cellcolor{green!20}{+3.5~\textsubscript{0.2}} & \cellcolor{green!20}{+1.2~\textsubscript{1.0}} \\
\cmidrule(lr){2-12}
 & \cellcolor{gray!20}OLMo3 (PT-Only) & \cellcolor{gray!20}27.5 & \cellcolor{gray!20}60.4 & \cellcolor{gray!20}58.9 & \cellcolor{gray!20}39.2 & \cellcolor{gray!20}26.0 & \cellcolor{gray!20}70.5 & \cellcolor{gray!20}64.4 & \cellcolor{gray!20}38.2 & \cellcolor{gray!20}35.2 & \cellcolor{gray!20}33.7 \\
 & Control & \cellcolor{green!20}{+1.2~\textsubscript{0.9}} & -0.0~\textsubscript{0.6} & \cellcolor{green!20}{+0.5~\textsubscript{1.8}} & -0.7~\textsubscript{2.2} & \cellcolor{green!20}{+1.0~\textsubscript{1.4}} & -1.7~\textsubscript{4.3} & \cellcolor{green!20}{+0.7~\textsubscript{0.6}} & -0.7~\textsubscript{1.1} & \cellcolor{green!20}{+0.6~\textsubscript{0.3}} & \cellcolor{green!20}{+0.5~\textsubscript{1.2}} \\
 & $k$-Shuffle Dyck & \cellcolor{green!20}{+0.5~\textsubscript{0.6}} & \cellcolor{green!20}{+0.3~\textsubscript{0.9}} & \cellcolor{green!20}{+0.6~\textsubscript{2.1}} & \cellcolor{green!20}{+0.7~\textsubscript{1.4}} & \cellcolor{green!20}{+0.2~\textsubscript{1.9}} & -2.8~\textsubscript{2.9} & \cellcolor{green!20}{+0.2~\textsubscript{0.7}} & -0.4~\textsubscript{0.6} & \cellcolor{green!20}{+1.1~\textsubscript{0.2}} & -0.2~\textsubscript{2.9} \\
\cmidrule(lr){2-12}
 & \cellcolor{gray!20}Marin (PT-Only) & \cellcolor{gray!20}33.7 & \cellcolor{gray!20}75.1 & \cellcolor{gray!20}75.9 & \cellcolor{gray!20}60.5 & \cellcolor{gray!20}32.0 & \cellcolor{gray!20}73.3 & \cellcolor{gray!20}69.8 & \cellcolor{gray!20}42.9 & \cellcolor{gray!20}47.4 & \cellcolor{gray!20}48.5 \\
 & Control & -0.7~\textsubscript{0.9} & \cellcolor{green!20}{+0.3~\textsubscript{0.3}} & \cellcolor{green!20}{+1.9~\textsubscript{2.2}} & \cellcolor{green!20}{+0.2~\textsubscript{0.6}} & \cellcolor{green!20}{+1.0~\textsubscript{0.6}} & -1.7~\textsubscript{3.2} & -0.0~\textsubscript{0.7} & -0.7~\textsubscript{0.5} & \cellcolor{green!20}{+0.3~\textsubscript{0.4}} & \cellcolor{green!20}{+0.6~\textsubscript{0.9}} \\
 & $k$-Shuffle Dyck & -0.6~\textsubscript{0.7} & \cellcolor{green!20}{+1.7~\textsubscript{0.5}} & \cellcolor{green!20}{+2.4~\textsubscript{1.3}} & \cellcolor{green!20}{+2.6~\textsubscript{0.7}} & \cellcolor{green!20}{+1.9~\textsubscript{1.4}} & \cellcolor{green!20}{+0.8~\textsubscript{2.9}} & \cellcolor{green!20}{+2.3~\textsubscript{0.6}} & \cellcolor{green!20}{+0.4~\textsubscript{0.6}} & \cellcolor{green!20}{+3.6~\textsubscript{0.2}} & \cellcolor{green!20}{+3.1~\textsubscript{1.3}} \\
\bottomrule
\end{tabular}
}
\end{center}
\end{table*}

\begin{table*}[!t]
\tiny
\caption{Linguistic competence on BLiMP by phenomenon. \colorbox{gray!20}{Gray} rows give the PT-Only mean over checkpoints at 5K--10K steps; the rows below give the mean paired difference from that baseline, with standard deviations across checkpoints as subscripts. \colorbox{green!20}{Green} marks an improvement.}
\label{tab:blimp_result_merged}
\setlength{\aboverulesep}{1.3pt}
\setlength{\belowrulesep}{1.3pt}
\renewcommand{\arraystretch}{1.0}
\setlength{\tabcolsep}{3pt}
\begin{center}
\resizebox{\textwidth}{!}{
\begin{tabular}{llccccccccccccc}
\toprule
 &  & \multicolumn{2}{c}{\textbf{Semantics}} & \multicolumn{4}{c}{\textbf{Morphology}} & \multicolumn{6}{c}{\textbf{Syntax}} & \multicolumn{1}{c}{\textbf{}} \\
\cmidrule(lr){3-4} \cmidrule(lr){5-8} \cmidrule(lr){9-14} \cmidrule(lr){15-15}
 & \textbf{Approach} & \textbf{Quant} & \textbf{NPI} & \textbf{Ana Agr} & \textbf{Irregul} & \textbf{DN Agr} & \textbf{SV Agr} & \textbf{Arg Str} & \textbf{Bind} & \textbf{Ctrl Rais} & \textbf{Ellips} & \textbf{Fill Gap} & \textbf{Island} & \textbf{Overall} \\
\midrule
\multirow{12}{*}{\rotatebox[origin=c]{90}{\textbf{500M}}} & \cellcolor{gray!20}C4 (PT-Only) & \cellcolor{gray!20}71.0 & \cellcolor{gray!20}64.5 & \cellcolor{gray!20}97.5 & \cellcolor{gray!20}93.9 & \cellcolor{gray!20}94.8 & \cellcolor{gray!20}85.8 & \cellcolor{gray!20}80.3 & \cellcolor{gray!20}77.2 & \cellcolor{gray!20}81.4 & \cellcolor{gray!20}87.2 & \cellcolor{gray!20}78.7 & \cellcolor{gray!20}68.8 & \cellcolor{gray!20}79.6 \\
 & Control & -1.3~\textsubscript{4.2} & -0.6~\textsubscript{9.3} & -1.3~\textsubscript{3.1} & \cellcolor{green!20}{+0.7~\textsubscript{2.0}} & -1.4~\textsubscript{1.3} & -2.0~\textsubscript{1.8} & -1.6~\textsubscript{2.0} & \cellcolor{green!20}{+2.2~\textsubscript{2.2}} & -2.2~\textsubscript{2.4} & -0.2~\textsubscript{1.6} & -1.7~\textsubscript{1.8} & -6.4~\textsubscript{2.2} & -1.6~\textsubscript{1.5} \\
 & $k$-Shuffle Dyck & \cellcolor{green!20}{+3.7~\textsubscript{5.9}} & -2.1~\textsubscript{3.9} & -2.4~\textsubscript{2.9} & -1.0~\textsubscript{1.6} & -0.4~\textsubscript{0.4} & -2.0~\textsubscript{3.2} & -0.5~\textsubscript{0.8} & \cellcolor{green!20}{+2.9~\textsubscript{1.2}} & -0.8~\textsubscript{1.0} & \cellcolor{green!20}{+0.2~\textsubscript{1.8}} & \cellcolor{green!20}{+0.6~\textsubscript{1.0}} & -3.9~\textsubscript{3.4} & -0.6~\textsubscript{1.0} \\
\cmidrule(lr){2-15}
 & \cellcolor{gray!20}SmolLM3 (PT-Only) & \cellcolor{gray!20}73.8 & \cellcolor{gray!20}57.0 & \cellcolor{gray!20}96.0 & \cellcolor{gray!20}94.8 & \cellcolor{gray!20}95.6 & \cellcolor{gray!20}86.7 & \cellcolor{gray!20}80.9 & \cellcolor{gray!20}79.7 & \cellcolor{gray!20}82.4 & \cellcolor{gray!20}88.3 & \cellcolor{gray!20}80.7 & \cellcolor{gray!20}67.7 & \cellcolor{gray!20}79.7 \\
 & Control & -3.2~\textsubscript{4.6} & \cellcolor{green!20}{+12.1~\textsubscript{7.7}} & \cellcolor{green!20}{+0.8~\textsubscript{3.3}} & \cellcolor{green!20}{+0.4~\textsubscript{2.0}} & -0.1~\textsubscript{1.1} & \cellcolor{green!20}{+1.3~\textsubscript{2.5}} & -0.1~\textsubscript{1.3} & \cellcolor{green!20}{+1.7~\textsubscript{1.2}} & -2.7~\textsubscript{1.0} & -0.1~\textsubscript{1.1} & \cellcolor{green!20}{+1.3~\textsubscript{1.0}} & -0.0~\textsubscript{3.0} & \cellcolor{green!20}{+1.3~\textsubscript{1.7}} \\
 & $k$-Shuffle Dyck & \cellcolor{green!20}{+0.4~\textsubscript{8.0}} & \cellcolor{green!20}{+0.3~\textsubscript{10.6}} & -1.3~\textsubscript{3.9} & -0.7~\textsubscript{2.2} & \cellcolor{green!20}{+0.6~\textsubscript{0.7}} & \cellcolor{green!20}{+0.4~\textsubscript{2.1}} & -0.9~\textsubscript{1.4} & \cellcolor{green!20}{+0.6~\textsubscript{1.7}} & +0.0~\textsubscript{0.6} & \cellcolor{green!20}{+0.4~\textsubscript{1.4}} & -2.7~\textsubscript{1.0} & -0.4~\textsubscript{2.2} & -0.3~\textsubscript{2.0} \\
\cmidrule(lr){2-15}
 & \cellcolor{gray!20}OLMo3 (PT-Only) & \cellcolor{gray!20}65.7 & \cellcolor{gray!20}56.8 & \cellcolor{gray!20}98.3 & \cellcolor{gray!20}94.0 & \cellcolor{gray!20}94.5 & \cellcolor{gray!20}83.5 & \cellcolor{gray!20}80.3 & \cellcolor{gray!20}79.2 & \cellcolor{gray!20}80.1 & \cellcolor{gray!20}88.3 & \cellcolor{gray!20}76.7 & \cellcolor{gray!20}61.0 & \cellcolor{gray!20}77.3 \\
 & Control & \cellcolor{green!20}{+4.6~\textsubscript{4.6}} & \cellcolor{green!20}{+2.1~\textsubscript{11.0}} & -0.7~\textsubscript{1.0} & -1.5~\textsubscript{2.0} & \cellcolor{green!20}{+0.1~\textsubscript{0.6}} & -0.2~\textsubscript{3.0} & -0.3~\textsubscript{1.4} & \cellcolor{green!20}{+2.3~\textsubscript{1.5}} & -0.5~\textsubscript{0.7} & -3.3~\textsubscript{2.3} & \cellcolor{green!20}{+1.1~\textsubscript{0.6}} & -3.5~\textsubscript{2.5} & \cellcolor{green!20}{+0.2~\textsubscript{1.2}} \\
 & $k$-Shuffle Dyck & \cellcolor{green!20}{+1.9~\textsubscript{6.5}} & \cellcolor{green!20}{+7.5~\textsubscript{4.6}} & -1.4~\textsubscript{2.0} & -2.5~\textsubscript{2.8} & -1.4~\textsubscript{0.6} & \cellcolor{green!20}{+2.2~\textsubscript{3.0}} & \cellcolor{green!20}{+0.4~\textsubscript{1.2}} & -2.0~\textsubscript{2.1} & \cellcolor{green!20}{+0.4~\textsubscript{0.2}} & -3.2~\textsubscript{1.3} & -1.3~\textsubscript{1.3} & \cellcolor{green!20}{+1.5~\textsubscript{2.8}} & \cellcolor{green!20}{+0.6~\textsubscript{1.0}} \\
\cmidrule(lr){2-15}
 & \cellcolor{gray!20}Marin (PT-Only) & \cellcolor{gray!20}70.4 & \cellcolor{gray!20}60.7 & \cellcolor{gray!20}97.9 & \cellcolor{gray!20}91.9 & \cellcolor{gray!20}95.8 & \cellcolor{gray!20}88.3 & \cellcolor{gray!20}81.2 & \cellcolor{gray!20}82.1 & \cellcolor{gray!20}79.4 & \cellcolor{gray!20}88.3 & \cellcolor{gray!20}80.2 & \cellcolor{gray!20}68.9 & \cellcolor{gray!20}80.2 \\
 & Control & \cellcolor{green!20}{+5.1~\textsubscript{5.0}} & \cellcolor{green!20}{+12.5~\textsubscript{8.2}} & \cellcolor{green!20}{+0.2~\textsubscript{0.7}} & \cellcolor{green!20}{+2.7~\textsubscript{2.2}} & +0.0~\textsubscript{0.5} & \cellcolor{green!20}{+0.9~\textsubscript{2.3}} & \cellcolor{green!20}{+0.4~\textsubscript{0.8}} & -1.8~\textsubscript{0.3} & \cellcolor{green!20}{+1.8~\textsubscript{0.8}} & \cellcolor{green!20}{+0.9~\textsubscript{0.5}} & -0.7~\textsubscript{0.7} & \cellcolor{green!20}{+1.7~\textsubscript{4.0}} & \cellcolor{green!20}{+1.9~\textsubscript{1.4}} \\
 & $k$-Shuffle Dyck & \cellcolor{green!20}{+0.1~\textsubscript{5.0}} & \cellcolor{green!20}{+2.3~\textsubscript{7.1}} & \cellcolor{green!20}{+1.0~\textsubscript{0.6}} & \cellcolor{green!20}{+3.3~\textsubscript{1.7}} & -0.0~\textsubscript{0.7} & -1.8~\textsubscript{2.0} & \cellcolor{green!20}{+0.2~\textsubscript{1.3}} & -0.1~\textsubscript{0.6} & \cellcolor{green!20}{+3.5~\textsubscript{0.9}} & \cellcolor{green!20}{+0.3~\textsubscript{1.6}} & -1.4~\textsubscript{0.5} & -0.8~\textsubscript{3.1} & \cellcolor{green!20}{+0.3~\textsubscript{1.4}} \\
\midrule
\multirow{12}{*}{\rotatebox[origin=c]{90}{\textbf{1B}}} & \cellcolor{gray!20}C4 (PT-Only) & \cellcolor{gray!20}79.2 & \cellcolor{gray!20}65.1 & \cellcolor{gray!20}96.5 & \cellcolor{gray!20}90.8 & \cellcolor{gray!20}93.9 & \cellcolor{gray!20}86.0 & \cellcolor{gray!20}79.1 & \cellcolor{gray!20}78.8 & \cellcolor{gray!20}80.3 & \cellcolor{gray!20}87.4 & \cellcolor{gray!20}78.8 & \cellcolor{gray!20}68.7 & \cellcolor{gray!20}79.9 \\
 & Control & -7.4~\textsubscript{7.1} & -4.7~\textsubscript{5.0} & \cellcolor{green!20}{+0.8~\textsubscript{1.2}} & \cellcolor{green!20}{+2.1~\textsubscript{1.4}} & \cellcolor{green!20}{+1.2~\textsubscript{0.8}} & \cellcolor{green!20}{+0.8~\textsubscript{2.4}} & \cellcolor{green!20}{+0.1~\textsubscript{0.7}} & \cellcolor{green!20}{+0.4~\textsubscript{0.9}} & \cellcolor{green!20}{+0.9~\textsubscript{0.9}} & \cellcolor{green!20}{+0.8~\textsubscript{0.8}} & \cellcolor{green!20}{+0.5~\textsubscript{1.6}} & -4.3~\textsubscript{3.3} & -1.0~\textsubscript{0.7} \\
 & $k$-Shuffle Dyck & -6.2~\textsubscript{5.4} & \cellcolor{green!20}{+5.1~\textsubscript{4.8}} & \cellcolor{green!20}{+1.5~\textsubscript{1.6}} & \cellcolor{green!20}{+3.8~\textsubscript{3.3}} & \cellcolor{green!20}{+0.7~\textsubscript{0.9}} & -0.7~\textsubscript{2.3} & \cellcolor{green!20}{+0.3~\textsubscript{1.3}} & \cellcolor{green!20}{+1.1~\textsubscript{1.0}} & -0.2~\textsubscript{0.6} & \cellcolor{green!20}{+0.6~\textsubscript{1.0}} & -0.3~\textsubscript{1.7} & -1.2~\textsubscript{2.7} & \cellcolor{green!20}{+0.3~\textsubscript{1.0}} \\
\cmidrule(lr){2-15}
 & \cellcolor{gray!20}SmolLM3 (PT-Only) & \cellcolor{gray!20}74.3 & \cellcolor{gray!20}63.1 & \cellcolor{gray!20}97.1 & \cellcolor{gray!20}95.1 & \cellcolor{gray!20}95.8 & \cellcolor{gray!20}89.5 & \cellcolor{gray!20}80.0 & \cellcolor{gray!20}81.4 & \cellcolor{gray!20}80.4 & \cellcolor{gray!20}87.8 & \cellcolor{gray!20}79.7 & \cellcolor{gray!20}69.3 & \cellcolor{gray!20}80.7 \\
 & Control & \cellcolor{green!20}{+1.6~\textsubscript{10.0}} & \cellcolor{green!20}{+2.4~\textsubscript{7.0}} & -6.4~\textsubscript{3.1} & -2.5~\textsubscript{2.3} & \cellcolor{green!20}{+0.3~\textsubscript{1.2}} & -2.8~\textsubscript{2.0} & \cellcolor{green!20}{+0.6~\textsubscript{1.0}} & \cellcolor{green!20}{+0.2~\textsubscript{1.6}} & -0.1~\textsubscript{1.2} & \cellcolor{green!20}{+1.4~\textsubscript{1.3}} & \cellcolor{green!20}{+0.9~\textsubscript{0.8}} & +0.0~\textsubscript{4.3} & \cellcolor{green!20}{+0.1~\textsubscript{1.4}} \\
 & $k$-Shuffle Dyck & -2.8~\textsubscript{7.9} & \cellcolor{green!20}{+7.2~\textsubscript{6.6}} & \cellcolor{green!20}{+1.2~\textsubscript{1.8}} & -0.3~\textsubscript{0.7} & -0.6~\textsubscript{0.8} & -2.6~\textsubscript{2.4} & \cellcolor{green!20}{+1.1~\textsubscript{1.0}} & -0.5~\textsubscript{1.7} & \cellcolor{green!20}{+1.6~\textsubscript{0.6}} & \cellcolor{green!20}{+0.6~\textsubscript{1.3}} & -0.0~\textsubscript{0.8} & \cellcolor{green!20}{+1.3~\textsubscript{2.7}} & \cellcolor{green!20}{+0.7~\textsubscript{1.2}} \\
\cmidrule(lr){2-15}
 & \cellcolor{gray!20}OLMo3 (PT-Only) & \cellcolor{gray!20}66.9 & \cellcolor{gray!20}54.3 & \cellcolor{gray!20}90.7 & \cellcolor{gray!20}86.7 & \cellcolor{gray!20}92.3 & \cellcolor{gray!20}76.1 & \cellcolor{gray!20}77.7 & \cellcolor{gray!20}76.4 & \cellcolor{gray!20}77.9 & \cellcolor{gray!20}81.9 & \cellcolor{gray!20}76.3 & \cellcolor{gray!20}57.1 & \cellcolor{gray!20}74.3 \\
 & Control & \cellcolor{green!20}{+10.6~\textsubscript{4.6}} & \cellcolor{green!20}{+5.2~\textsubscript{10.2}} & \cellcolor{green!20}{+5.9~\textsubscript{3.5}} & \cellcolor{green!20}{+4.3~\textsubscript{6.8}} & \cellcolor{green!20}{+1.9~\textsubscript{2.0}} & \cellcolor{green!20}{+7.1~\textsubscript{8.0}} & \cellcolor{green!20}{+1.9~\textsubscript{1.8}} & \cellcolor{green!20}{+2.3~\textsubscript{1.5}} & \cellcolor{green!20}{+0.8~\textsubscript{1.9}} & \cellcolor{green!20}{+3.2~\textsubscript{2.6}} & \cellcolor{green!20}{+2.4~\textsubscript{1.8}} & \cellcolor{green!20}{+3.7~\textsubscript{3.9}} & \cellcolor{green!20}{+3.7~\textsubscript{2.5}} \\
 & $k$-Shuffle Dyck & \cellcolor{green!20}{+8.1~\textsubscript{4.3}} & \cellcolor{green!20}{+9.8~\textsubscript{7.8}} & \cellcolor{green!20}{+6.2~\textsubscript{3.4}} & \cellcolor{green!20}{+6.5~\textsubscript{4.4}} & \cellcolor{green!20}{+1.2~\textsubscript{2.1}} & \cellcolor{green!20}{+8.1~\textsubscript{5.3}} & \cellcolor{green!20}{+2.4~\textsubscript{1.8}} & \cellcolor{green!20}{+2.6~\textsubscript{1.3}} & \cellcolor{green!20}{+1.9~\textsubscript{2.5}} & \cellcolor{green!20}{+3.5~\textsubscript{2.5}} & \cellcolor{green!20}{+0.7~\textsubscript{1.7}} & \cellcolor{green!20}{+3.7~\textsubscript{4.0}} & \cellcolor{green!20}{+4.1~\textsubscript{1.9}} \\
\cmidrule(lr){2-15}
 & \cellcolor{gray!20}Marin (PT-Only) & \cellcolor{gray!20}75.6 & \cellcolor{gray!20}58.3 & \cellcolor{gray!20}96.9 & \cellcolor{gray!20}92.8 & \cellcolor{gray!20}95.9 & \cellcolor{gray!20}86.8 & \cellcolor{gray!20}80.5 & \cellcolor{gray!20}81.4 & \cellcolor{gray!20}80.8 & \cellcolor{gray!20}89.5 & \cellcolor{gray!20}78.9 & \cellcolor{gray!20}67.5 & \cellcolor{gray!20}79.8 \\
 & Control & -4.8~\textsubscript{9.6} & \cellcolor{green!20}{+1.7~\textsubscript{13.4}} & -2.6~\textsubscript{4.0} & -1.2~\textsubscript{3.3} & -0.1~\textsubscript{0.4} & -0.1~\textsubscript{2.6} & -1.5~\textsubscript{1.2} & \cellcolor{green!20}{+0.9~\textsubscript{0.5}} & \cellcolor{green!20}{+0.2~\textsubscript{0.6}} & \cellcolor{green!20}{+0.3~\textsubscript{1.2}} & \cellcolor{green!20}{+1.7~\textsubscript{0.9}} & \cellcolor{green!20}{+1.0~\textsubscript{2.7}} & -0.0~\textsubscript{2.5} \\
 & $k$-Shuffle Dyck & -2.0~\textsubscript{3.8} & \cellcolor{green!20}{+10.5~\textsubscript{11.6}} & \cellcolor{green!20}{+2.1~\textsubscript{2.2}} & -0.1~\textsubscript{4.4} & -0.9~\textsubscript{0.8} & \cellcolor{green!20}{+1.4~\textsubscript{2.3}} & \cellcolor{green!20}{+0.6~\textsubscript{1.9}} & -0.2~\textsubscript{1.2} & \cellcolor{green!20}{+1.0~\textsubscript{0.8}} & -0.7~\textsubscript{1.0} & \cellcolor{green!20}{+0.4~\textsubscript{1.1}} & \cellcolor{green!20}{+3.4~\textsubscript{3.0}} & \cellcolor{green!20}{+1.6~\textsubscript{1.8}} \\
\midrule
\multirow{12}{*}{\rotatebox[origin=c]{90}{\textbf{3B}}} & \cellcolor{gray!20}C4 (PT-Only) & \cellcolor{gray!20}68.4 & \cellcolor{gray!20}59.9 & \cellcolor{gray!20}94.4 & \cellcolor{gray!20}91.7 & \cellcolor{gray!20}95.5 & \cellcolor{gray!20}82.4 & \cellcolor{gray!20}79.0 & \cellcolor{gray!20}81.3 & \cellcolor{gray!20}80.1 & \cellcolor{gray!20}89.0 & \cellcolor{gray!20}78.6 & \cellcolor{gray!20}64.3 & \cellcolor{gray!20}78.3 \\
 & Control & \cellcolor{green!20}{+4.5~\textsubscript{7.9}} & \cellcolor{green!20}{+4.4~\textsubscript{11.8}} & -2.6~\textsubscript{5.8} & -1.5~\textsubscript{8.1} & -1.0~\textsubscript{1.2} & \cellcolor{green!20}{+2.2~\textsubscript{3.8}} & -1.0~\textsubscript{1.3} & \cellcolor{green!20}{+0.9~\textsubscript{0.7}} & -0.1~\textsubscript{0.4} & \cellcolor{green!20}{+0.6~\textsubscript{1.1}} & -1.6~\textsubscript{1.6} & -0.8~\textsubscript{5.7} & \cellcolor{green!20}{+0.4~\textsubscript{2.5}} \\
 & $k$-Shuffle Dyck & \cellcolor{green!20}{+12.5~\textsubscript{8.7}} & -1.0~\textsubscript{8.8} & \cellcolor{green!20}{+4.2~\textsubscript{2.9}} & \cellcolor{green!20}{+1.9~\textsubscript{6.2}} & -0.9~\textsubscript{0.6} & \cellcolor{green!20}{+3.3~\textsubscript{3.8}} & \cellcolor{green!20}{+1.1~\textsubscript{1.4}} & -0.9~\textsubscript{0.8} & \cellcolor{green!20}{+1.1~\textsubscript{0.8}} & \cellcolor{green!20}{+0.1~\textsubscript{1.4}} & -0.8~\textsubscript{0.8} & \cellcolor{green!20}{+5.1~\textsubscript{5.9}} & \cellcolor{green!20}{+1.7~\textsubscript{1.8}} \\
\cmidrule(lr){2-15}
 & \cellcolor{gray!20}SmolLM3 (PT-Only) & \cellcolor{gray!20}74.6 & \cellcolor{gray!20}64.7 & \cellcolor{gray!20}96.6 & \cellcolor{gray!20}93.9 & \cellcolor{gray!20}95.6 & \cellcolor{gray!20}85.1 & \cellcolor{gray!20}79.2 & \cellcolor{gray!20}79.7 & \cellcolor{gray!20}79.8 & \cellcolor{gray!20}88.7 & \cellcolor{gray!20}78.8 & \cellcolor{gray!20}72.0 & \cellcolor{gray!20}80.3 \\
 & Control & \cellcolor{green!20}{+2.8~\textsubscript{4.5}} & \cellcolor{green!20}{+6.0~\textsubscript{5.7}} & \cellcolor{green!20}{+2.4~\textsubscript{1.2}} & -1.5~\textsubscript{3.2} & \cellcolor{green!20}{+1.0~\textsubscript{0.6}} & \cellcolor{green!20}{+4.6~\textsubscript{1.2}} & \cellcolor{green!20}{+2.1~\textsubscript{0.5}} & \cellcolor{green!20}{+2.0~\textsubscript{1.3}} & \cellcolor{green!20}{+1.7~\textsubscript{0.6}} & \cellcolor{green!20}{+1.0~\textsubscript{1.2}} & \cellcolor{green!20}{+1.2~\textsubscript{1.2}} & -0.1~\textsubscript{1.1} & \cellcolor{green!20}{+2.1~\textsubscript{1.0}} \\
 & $k$-Shuffle Dyck & \cellcolor{green!20}{+1.0~\textsubscript{4.9}} & -0.5~\textsubscript{9.8} & \cellcolor{green!20}{+1.6~\textsubscript{0.7}} & -0.7~\textsubscript{3.0} & \cellcolor{green!20}{+0.6~\textsubscript{0.8}} & \cellcolor{green!20}{+1.7~\textsubscript{2.5}} & -0.2~\textsubscript{1.6} & \cellcolor{green!20}{+2.6~\textsubscript{1.6}} & \cellcolor{green!20}{+2.1~\textsubscript{1.5}} & -0.1~\textsubscript{1.3} & \cellcolor{green!20}{+0.7~\textsubscript{1.2}} & -0.4~\textsubscript{1.9} & \cellcolor{green!20}{+0.7~\textsubscript{1.9}} \\
\cmidrule(lr){2-15}
 & \cellcolor{gray!20}OLMo3 (PT-Only) & \cellcolor{gray!20}75.7 & \cellcolor{gray!20}62.0 & \cellcolor{gray!20}95.3 & \cellcolor{gray!20}87.2 & \cellcolor{gray!20}93.5 & \cellcolor{gray!20}82.0 & \cellcolor{gray!20}79.6 & \cellcolor{gray!20}80.5 & \cellcolor{gray!20}76.1 & \cellcolor{gray!20}85.6 & \cellcolor{gray!20}76.5 & \cellcolor{gray!20}61.0 & \cellcolor{gray!20}77.6 \\
 & Control & -5.0~\textsubscript{7.9} & \cellcolor{green!20}{+4.2~\textsubscript{5.7}} & \cellcolor{green!20}{+2.7~\textsubscript{1.8}} & \cellcolor{green!20}{+3.0~\textsubscript{6.1}} & -0.7~\textsubscript{0.9} & -1.8~\textsubscript{3.3} & -0.1~\textsubscript{0.9} & -0.5~\textsubscript{2.0} & -0.1~\textsubscript{6.7} & -0.9~\textsubscript{1.2} & -0.7~\textsubscript{1.2} & -0.6~\textsubscript{2.5} & -0.2~\textsubscript{1.6} \\
 & $k$-Shuffle Dyck & -1.3~\textsubscript{7.5} & -6.4~\textsubscript{7.4} & \cellcolor{green!20}{+1.9~\textsubscript{2.5}} & \cellcolor{green!20}{+4.2~\textsubscript{6.7}} & \cellcolor{green!20}{+0.1~\textsubscript{1.3}} & \cellcolor{green!20}{+0.6~\textsubscript{3.1}} & \cellcolor{green!20}{+0.6~\textsubscript{0.8}} & -0.2~\textsubscript{2.6} & \cellcolor{green!20}{+2.9~\textsubscript{2.2}} & -0.9~\textsubscript{1.4} & \cellcolor{green!20}{+2.4~\textsubscript{1.7}} & \cellcolor{green!20}{+0.4~\textsubscript{3.8}} & \cellcolor{green!20}{+0.1~\textsubscript{1.7}} \\
\cmidrule(lr){2-15}
 & \cellcolor{gray!20}Marin (PT-Only) & \cellcolor{gray!20}77.8 & \cellcolor{gray!20}58.5 & \cellcolor{gray!20}93.1 & \cellcolor{gray!20}88.9 & \cellcolor{gray!20}96.0 & \cellcolor{gray!20}86.2 & \cellcolor{gray!20}80.7 & \cellcolor{gray!20}82.2 & \cellcolor{gray!20}79.4 & \cellcolor{gray!20}89.5 & \cellcolor{gray!20}77.9 & \cellcolor{gray!20}68.2 & \cellcolor{gray!20}79.7 \\
 & Control & \cellcolor{green!20}{+0.2~\textsubscript{2.1}} & \cellcolor{green!20}{+1.4~\textsubscript{9.6}} & \cellcolor{green!20}{+5.1~\textsubscript{4.6}} & \cellcolor{green!20}{+5.5~\textsubscript{4.8}} & -1.9~\textsubscript{3.7} & \cellcolor{green!20}{+1.2~\textsubscript{3.4}} & -0.6~\textsubscript{1.3} & \cellcolor{green!20}{+0.1~\textsubscript{1.4}} & \cellcolor{green!20}{+0.8~\textsubscript{1.7}} & -2.4~\textsubscript{2.2} & -0.4~\textsubscript{0.6} & \cellcolor{green!20}{+3.0~\textsubscript{2.9}} & \cellcolor{green!20}{+0.6~\textsubscript{1.1}} \\
 & $k$-Shuffle Dyck & -8.7~\textsubscript{5.6} & \cellcolor{green!20}{+6.5~\textsubscript{6.8}} & \cellcolor{green!20}{+2.9~\textsubscript{3.2}} & \cellcolor{green!20}{+1.0~\textsubscript{3.8}} & -0.6~\textsubscript{0.9} & -0.1~\textsubscript{2.8} & -2.0~\textsubscript{1.3} & -1.9~\textsubscript{1.1} & \cellcolor{green!20}{+1.1~\textsubscript{1.6}} & -2.3~\textsubscript{0.4} & -0.5~\textsubscript{1.0} & \cellcolor{green!20}{+0.3~\textsubscript{3.2}} & -0.3~\textsubscript{1.0} \\
\bottomrule
\end{tabular}
}
\end{center}
\end{table*}

\begin{itemize}[leftmargin=*]
\item Table \ref{tab:appendix_section5_downstream_wide} shows task-level downstream score changes from PPT relative to PT-Only.

\item Table \ref{tab:blimp_result_merged} shows the breakdown of BLiMP score changes from PPT relative to PT-Only. In Morphology and Syntax, the delta of $k$-Shuffle Dyck stays at or below 1.0 point in 19 of the 24 subgroup cases, and is not stable in 20 of them.
Both movement and variance instead concentrate in Semantics, where the deciding evidence is lexical and pragmatic rather than structural: the average checkpoint SD for $k$-Shuffle Dyck is 5.7 in Semantics, against 1.5 in Morphology and 0.9 in Syntax.

\end{itemize}

\subsection{Checkpoint Sensitivity of Single-Snapshot Estimates}
\label{app:estimator}

\paragraph{Single Checkpoints versus Multi-Checkpoint Averages.}
Table~\ref{tab:estimator_audit} compares the single-checkpoint delta at step 10K against the multi-checkpoint average across steps 5K to 10K for $k$-Shuffle Dyck over all 12 scale-mixture pairs.
Downstream task averages demonstrate strong directional agreement. No pair reverses sign, and the two estimates differ by an average of only 0.54 points.
In contrast, BLiMP evaluations diverge substantially.
Half of the 12 pairs reverse sign, and the two estimates differ by 1.50 points on average.
On SmolLM3 at the 500M parameter scale, for example, a gain of 2.93 points at step 10K corresponds to a multi-checkpoint average of $-$0.28 points, with only two of six checkpoints favoring PPT.
The BLiMP differences reported in this paper are typically a few points, which falls below the intra-run variance observed across checkpoints.
A single checkpoint therefore cannot determine whether an observed margin on BLiMP reflects a true effect, which is why we average across checkpoints throughout.

\begin{table}[th]
\small
\caption{Single-checkpoint versus multi-checkpoint estimates of the $k$-Shuffle Dyck effect.
$\Delta_{10\text{K}}$ denotes the paired difference at step 10K, while $\bar{\Delta}$ and \textbf{sd} represent the mean and standard deviation across checkpoints from 5K to 10K steps.
The column \emph{Ckpts} reports the ratio of checkpoints favoring PPT, and $\dagger$ marks configurations with directional disagreement between the two estimates.}
\label{tab:estimator_audit}
\begin{center}
\begin{tabular}{llrrrc}
\toprule
\textbf{Scale} & \textbf{Data} & $\Delta_{10\text{K}}$ & $\bar{\Delta}$ & \textbf{sd} & \textbf{Ckpts} \\
\midrule
\multicolumn{6}{l}{\textit{Downstream Average}} \\
500M & C4 & +0.66 & +0.28 & 0.23 & 6/6 \\
500M & Marin & +1.89 & +1.39 & 0.57 & 6/6 \\
500M & OLMo3 & +1.38 & +0.66 & 0.62 & 4/6 \\
500M & SmolLM3 & +2.64 & +0.98 & 0.94 & 5/6 \\
1B & C4 & +2.94 & +2.23 & 0.73 & 6/6 \\
1B & Marin & +2.23 & +2.00 & 0.33 & 6/6 \\
1B & OLMo3 & -0.40 & -0.47 & 0.68 & 1/6 \\
1B & SmolLM3 & +1.63 & +1.72 & 0.47 & 6/6 \\
3B & C4 & +1.94 & +1.32 & 0.55 & 6/6 \\
3B & Marin & +1.45 & +1.95 & 0.54 & 6/6 \\
3B & OLMo3 & +0.65 & +0.04 & 1.21 & 3/6 \\
3B & SmolLM3 & +2.09 & +1.75 & 0.35 & 6/6 \\
\midrule
\multicolumn{6}{l}{\textit{BLiMP Overall}} \\
500M & C4 & +1.05 & -0.56$^{\dagger}$ & 0.99 & 1/6 \\
500M & Marin & -1.07 & +0.26$^{\dagger}$ & 1.36 & 3/6 \\
500M & OLMo3 & +0.88 & +0.64 & 1.02 & 3/6 \\
500M & SmolLM3 & +2.93 & -0.28$^{\dagger}$ & 2.02 & 2/6 \\
1B & C4 & +1.73 & +0.34 & 1.03 & 3/6 \\
1B & Marin & +3.22 & +1.61 & 1.76 & 5/6 \\
1B & OLMo3 & +1.31 & +4.10 & 1.91 & 6/6 \\
1B & SmolLM3 & -0.32 & +0.71$^{\dagger}$ & 1.19 & 3/6 \\
3B & C4 & +3.76 & +1.66 & 1.77 & 5/6 \\
3B & Marin & +0.19 & -0.27$^{\dagger}$ & 0.97 & 3/6 \\
3B & OLMo3 & -1.96 & +0.05$^{\dagger}$ & 1.67 & 2/6 \\
3B & SmolLM3 & +0.47 & +0.67 & 1.93 & 4/6 \\
\bottomrule
\end{tabular}
\end{center}
\end{table}

\paragraph{Sensitivity to the Averaging Window.}
We average metrics across the second half of training, and verify that no conclusion depends on that choice.
Table~\ref{tab:window_sweep} sweeps the start of the window from 2K to 7K steps.
Across every starting checkpoint, downstream performance improves in 11 of the 12 pairs, verbatim retrieval improves in all 12, and no downstream pair reverses sign relative to the 10K snapshot.
In contrast, BLiMP scores reverse sign in five or six of the 12 pairs across the same windows.
Mean downstream gains shift by at most 0.3 points, preserving the relative ranking across model scales.
These findings therefore do not depend on where the window starts.

The default evaluation window spans both the steady learning rate interval and the final 1K cooling steps of the Warmup-Stable-Decay (WSD) schedule.
To verify that the observed advantages do not depend on this terminal phase, we also examine an alternative window restricted to steps 5K to 9K (Table~\ref{tab:window_sweep}).
Excluding the decay phase weakens the downstream estimate only slightly: 10 of 12 pairs improve rather than 11, and the 500M mean falls from $+$0.83 to $+$0.64.
All three scale means remain positive and their ordering is unchanged, as are the BLiMP and verbatim counts.
The reported advantages therefore persist independently of the terminal cooling phase.

\begin{table}[t]
\small
\caption{Sensitivity of downstream gains and linguistic benchmarks to the choice of averaging window. The metric \emph{Cells~$>0$} records the proportion of configurations where $k$-Shuffle Dyck yields improvements, whereas \emph{Flips} denotes directional disagreement relative to the 10K snapshot across all 12 scale and mixture combinations. \textbf{Bold} text denotes the baseline 5K--10K window adopted throughout this study.}
\label{tab:window_sweep}
\begin{center}
\begin{tabular}{lccccccc}
\toprule
& \multicolumn{5}{c}{\textbf{Downstream average}} & \textbf{BLiMP} & \textbf{Verbatim} \\
\cmidrule(lr){2-6}\cmidrule(lr){7-7}\cmidrule(lr){8-8}
\textbf{Window} & Cells $>0$ & Flips & 500M & 1B & 3B & Flips & Cells $>0$ \\
\midrule
2K--10K & 11/12 & 0/12 & +0.86 & +1.13 & +1.23 & 5/12 & 12/12 \\
3K--10K & 11/12 & 0/12 & +0.84 & +1.29 & +1.22 & 5/12 & 12/12 \\
4K--10K & 11/12 & 0/12 & +0.84 & +1.36 & +1.22 & 5/12 & 12/12 \\
\textbf{5K--10K} & 11/12 & 0/12 & +0.83 & +1.37 & +1.27 & 6/12 & 12/12 \\
6K--10K & 11/12 & 0/12 & +0.80 & +1.43 & +1.29 & 6/12 & 12/12 \\
7K--10K & 11/12 & 0/12 & +0.84 & +1.42 & +1.28 & 6/12 & 12/12 \\
5K--9K & 10/12 & 2/12 & +0.64 & +1.32 & +1.21 & 5/12 & 12/12 \\
\bottomrule
\end{tabular}
\end{center}
\end{table}

\clearpage
\subsection{Sensitivity to Random Seeds}
\label{app:seed}

\begin{table*}[!t]
\caption{Task-level downstream performance across random seeds (3B Marin). \colorbox{gray!20}{Gray} rows give the PT-Only mean over checkpoints at 5K--10K steps; the rows below give the mean paired difference from that baseline, with standard deviations across checkpoints as subscripts. \colorbox{green!20}{Green} marks an improvement.}
\label{tab:appendix_section5_downstream_multiseed_3b}
\begin{center}
\renewcommand{\arraystretch}{1.0}
\setlength{\tabcolsep}{3pt}
\resizebox{\textwidth}{!}{
\begin{tabular}{llcccccccccc}
\toprule
 &  & \multicolumn{2}{c}{\textbf{RC}} & \multicolumn{3}{c}{\textbf{Science QA}} & \multicolumn{4}{c}{\textbf{CR}} & \multicolumn{1}{c}{\textbf{LM}} \\
\cmidrule(lr){3-4} \cmidrule(lr){5-7} \cmidrule(lr){8-11} \cmidrule(lr){12-12}
 & \textbf{Approach} & \textbf{RACE} & \textbf{ReCoRD} & \textbf{SciQ} & \textbf{ARC-Easy} & \textbf{OpenBookQA} & \textbf{COPA} & \textbf{PIQA} & \textbf{SIQA} & \textbf{HellaSwag} & \textbf{LAMBADA} \\
\midrule
\multirow{2}{*}{\textbf{Seed 1}} & \cellcolor{gray!20}Marin 3B (PT-Only) & \cellcolor{gray!20}33.7 & \cellcolor{gray!20}75.1 & \cellcolor{gray!20}75.9 & \cellcolor{gray!20}60.5 & \cellcolor{gray!20}32.0 & \cellcolor{gray!20}73.3 & \cellcolor{gray!20}69.8 & \cellcolor{gray!20}42.9 & \cellcolor{gray!20}47.4 & \cellcolor{gray!20}48.5 \\
 & $k$-Shuffle Dyck & -0.6~\textsubscript{0.7} & \cellcolor{green!20}{+1.7~\textsubscript{0.5}} & \cellcolor{green!20}{+2.4~\textsubscript{1.3}} & \cellcolor{green!20}{+2.6~\textsubscript{0.7}} & \cellcolor{green!20}{+1.9~\textsubscript{1.4}} & \cellcolor{green!20}{+0.8~\textsubscript{2.9}} & \cellcolor{green!20}{+2.3~\textsubscript{0.6}} & \cellcolor{green!20}{+0.4~\textsubscript{0.6}} & \cellcolor{green!20}{+3.6~\textsubscript{0.2}} & \cellcolor{green!20}{+3.1~\textsubscript{1.3}} \\
\midrule
\multirow{2}{*}{\textbf{Seed 2}} & \cellcolor{gray!20}Marin 3B (PT-Only) & \cellcolor{gray!20}33.0 & \cellcolor{gray!20}74.6 & \cellcolor{gray!20}74.9 & \cellcolor{gray!20}60.5 & \cellcolor{gray!20}33.4 & \cellcolor{gray!20}71.7 & \cellcolor{gray!20}69.7 & \cellcolor{gray!20}42.4 & \cellcolor{gray!20}47.0 & \cellcolor{gray!20}48.1 \\
 & $k$-Shuffle Dyck & -0.4~\textsubscript{0.9} & \cellcolor{green!20}{+1.7~\textsubscript{1.0}} & \cellcolor{green!20}{+3.6~\textsubscript{1.7}} & \cellcolor{green!20}{+2.1~\textsubscript{0.4}} & \cellcolor{green!20}{+0.2~\textsubscript{0.9}} & \cellcolor{green!20}{+1.2~\textsubscript{3.7}} & \cellcolor{green!20}{+1.4~\textsubscript{0.9}} & \cellcolor{green!20}{+1.1~\textsubscript{0.8}} & \cellcolor{green!20}{+3.4~\textsubscript{0.3}} & \cellcolor{green!20}{+3.3~\textsubscript{1.3}} \\

\midrule
\multirow{2}{*}{\textbf{Seed 3}} & \cellcolor{gray!20}Marin 3B (PT-Only) & \cellcolor{gray!20}32.4 & \cellcolor{gray!20}75.2 & \cellcolor{gray!20}75.4 & \cellcolor{gray!20}61.3 & \cellcolor{gray!20}34.0 & \cellcolor{gray!20}72.8 & \cellcolor{gray!20}70.0 & \cellcolor{gray!20}41.4 & \cellcolor{gray!20}47.7 & \cellcolor{gray!20}49.4 \\
 & $k$-Shuffle Dyck & \cellcolor{green!20}{+0.6~\textsubscript{0.4}} & \cellcolor{green!20}{+0.9~\textsubscript{0.3}} & \cellcolor{green!20}{+2.3~\textsubscript{0.9}} & \cellcolor{green!20}{+1.6~\textsubscript{0.7}} & +0.0~\textsubscript{1.5} & \cellcolor{green!20}{+2.2~\textsubscript{1.5}} & \cellcolor{green!20}{+1.3~\textsubscript{0.6}} & \cellcolor{green!20}{+1.7~\textsubscript{0.5}} & \cellcolor{green!20}{+2.7~\textsubscript{0.3}} & \cellcolor{green!20}{+2.8~\textsubscript{1.1}} \\

\midrule
\multirow{2}{*}{\textbf{Seed Mean}} & \cellcolor{gray!20}Marin 3B (PT-Only) & \cellcolor{gray!20}33.0~\textsubscript{0.6} & \cellcolor{gray!20}75.0~\textsubscript{0.3} & \cellcolor{gray!20}75.4~\textsubscript{0.5} & \cellcolor{gray!20}60.8~\textsubscript{0.5} & \cellcolor{gray!20}33.1~\textsubscript{1.0} & \cellcolor{gray!20}72.6~\textsubscript{0.9} & \cellcolor{gray!20}69.8~\textsubscript{0.2} & \cellcolor{gray!20}42.3~\textsubscript{0.8} & \cellcolor{gray!20}47.4~\textsubscript{0.3} & \cellcolor{gray!20}48.7~\textsubscript{0.7} \\
 & $k$-Shuffle Dyck & -0.1~\textsubscript{0.6} & \cellcolor{green!20}{+1.4~\textsubscript{0.5}} & \cellcolor{green!20}{+2.8~\textsubscript{0.7}} & \cellcolor{green!20}{+2.1~\textsubscript{0.5}} & \cellcolor{green!20}{+0.7~\textsubscript{1.0}} & \cellcolor{green!20}{+1.4~\textsubscript{0.7}} & \cellcolor{green!20}{+1.7~\textsubscript{0.5}} & \cellcolor{green!20}{+1.1~\textsubscript{0.7}} & \cellcolor{green!20}{+3.2~\textsubscript{0.4}} & \cellcolor{green!20}{+3.1~\textsubscript{0.2}} \\

\bottomrule
\end{tabular}}
\end{center}
\end{table*}

\begin{table*}[!t]
\caption{Linguistic competence on BLiMP across random seeds (3B Marin). \colorbox{gray!20}{Gray} rows give the PT-Only mean over checkpoints at 5K--10K steps; the rows below give the mean paired difference from that baseline, with standard deviations across checkpoints as subscripts. \colorbox{green!20}{Green} marks an improvement.}
\label{tab:appendix_section5_blimp_multiseed_3b}
\renewcommand{\arraystretch}{1.0}
\setlength{\tabcolsep}{3pt}
\begin{center}
\resizebox{\textwidth}{!}{
\begin{tabular}{llccccccccccccc}
\toprule
 &  & \multicolumn{2}{c}{\textbf{Semantics}} & \multicolumn{4}{c}{\textbf{Morphology}} & \multicolumn{6}{c}{\textbf{Syntax}} & \multicolumn{1}{c}{} \\
\cmidrule(lr){3-4} \cmidrule(lr){5-8} \cmidrule(lr){9-14}
 & \textbf{Approach} & \textbf{Quant} & \textbf{NPI} & \textbf{Ana Agr} & \textbf{Irregul} & \textbf{DN Agr} & \textbf{SV Agr} & \textbf{Arg Str} & \textbf{Bind} & \textbf{Ctrl Rais} & \textbf{Ellips} & \textbf{Fill Gap} & \textbf{Island} & \textbf{Overall} \\
\midrule
\multirow{2}{*}{\textbf{Seed 1}} & \cellcolor{gray!20}Marin 3B (PT-Only) & \cellcolor{gray!20}77.8 & \cellcolor{gray!20}58.5 & \cellcolor{gray!20}93.1 & \cellcolor{gray!20}88.9 & \cellcolor{gray!20}96.0 & \cellcolor{gray!20}86.2 & \cellcolor{gray!20}80.7 & \cellcolor{gray!20}82.2 & \cellcolor{gray!20}79.4 & \cellcolor{gray!20}89.5 & \cellcolor{gray!20}77.9 & \cellcolor{gray!20}68.2 & \cellcolor{gray!20}79.7 \\
 & $k$-Shuffle Dyck & -8.7~\textsubscript{5.6} & \cellcolor{green!20}{+6.5~\textsubscript{6.8}} & \cellcolor{green!20}{+2.9~\textsubscript{3.2}} & \cellcolor{green!20}{+1.0~\textsubscript{3.8}} & -0.6~\textsubscript{0.9} & -0.1~\textsubscript{2.8} & -2.0~\textsubscript{1.3} & -1.9~\textsubscript{1.1} & \cellcolor{green!20}{+1.1~\textsubscript{1.6}} & -2.3~\textsubscript{0.4} & -0.5~\textsubscript{1.0} & \cellcolor{green!20}{+0.3~\textsubscript{3.2}} & -0.3~\textsubscript{1.0} \\
\midrule
\multirow{2}{*}{\textbf{Seed 2}} & \cellcolor{gray!20}Marin 3B (PT-Only) & \cellcolor{gray!20}72.4 & \cellcolor{gray!20}66.1 & \cellcolor{gray!20}90.9 & \cellcolor{gray!20}90.0 & \cellcolor{gray!20}95.3 & \cellcolor{gray!20}87.2 & \cellcolor{gray!20}80.0 & \cellcolor{gray!20}82.3 & \cellcolor{gray!20}80.4 & \cellcolor{gray!20}89.6 & \cellcolor{gray!20}78.7 & \cellcolor{gray!20}68.3 & \cellcolor{gray!20}80.2 \\
 & $k$-Shuffle Dyck & \cellcolor{green!20}{+3.9~\textsubscript{9.3}} & -3.0~\textsubscript{8.2} & \cellcolor{green!20}{+7.9~\textsubscript{6.8}} & \cellcolor{green!20}{+4.0~\textsubscript{6.4}} & -0.2~\textsubscript{0.4} & -1.5~\textsubscript{2.4} & -0.7~\textsubscript{1.3} & -0.3~\textsubscript{0.9} & \cellcolor{green!20}{+0.5~\textsubscript{1.0}} & -1.9~\textsubscript{1.0} & -0.4~\textsubscript{1.3} & \cellcolor{green!20}{+1.5~\textsubscript{3.3}} & \cellcolor{green!20}{+0.1~\textsubscript{1.9}} \\
\midrule
\multirow{2}{*}{\textbf{Seed 3}} & \cellcolor{gray!20}Marin 3B (PT-Only) & \cellcolor{gray!20}79.3 & \cellcolor{gray!20}66.5 & \cellcolor{gray!20}94.1 & \cellcolor{gray!20}87.7 & \cellcolor{gray!20}96.2 & \cellcolor{gray!20}85.5 & \cellcolor{gray!20}81.8 & \cellcolor{gray!20}82.2 & \cellcolor{gray!20}80.2 & \cellcolor{gray!20}88.9 & \cellcolor{gray!20}79.3 & \cellcolor{gray!20}71.6 & \cellcolor{gray!20}81.3 \\
 & $k$-Shuffle Dyck & -0.3~\textsubscript{6.7} & \cellcolor{green!20}{+3.9~\textsubscript{5.3}} & \cellcolor{green!20}{+4.7~\textsubscript{3.8}} & \cellcolor{green!20}{+5.9~\textsubscript{1.4}} & -1.1~\textsubscript{1.1} & \cellcolor{green!20}{+2.8~\textsubscript{2.8}} & -1.3~\textsubscript{0.7} & \cellcolor{green!20}{+0.7~\textsubscript{1.0}} & \cellcolor{green!20}{+1.9~\textsubscript{1.3}} & -1.3~\textsubscript{2.0} & -1.4~\textsubscript{1.0} & \cellcolor{green!20}{+1.9~\textsubscript{4.2}} & \cellcolor{green!20}{+0.9~\textsubscript{1.3}} \\
\midrule
\multirow{2}{*}{\textbf{Seed Mean}} & \cellcolor{gray!20}Marin 3B (PT-Only) & \cellcolor{gray!20}76.5~\textsubscript{3.6} & \cellcolor{gray!20}63.7~\textsubscript{4.5} & \cellcolor{gray!20}92.7~\textsubscript{1.6} & \cellcolor{gray!20}88.9~\textsubscript{1.1} & \cellcolor{gray!20}95.8~\textsubscript{0.5} & \cellcolor{gray!20}86.3~\textsubscript{0.9} & \cellcolor{gray!20}80.8~\textsubscript{0.9} & \cellcolor{gray!20}82.3~\textsubscript{0.1} & \cellcolor{gray!20}80.0~\textsubscript{0.5} & \cellcolor{gray!20}89.3~\textsubscript{0.4} & \cellcolor{gray!20}78.6~\textsubscript{0.7} & \cellcolor{gray!20}69.4~\textsubscript{1.9} & \cellcolor{gray!20}80.4~\textsubscript{0.8} \\
 & $k$-Shuffle Dyck & -1.7~\textsubscript{6.4} & \cellcolor{green!20}{+2.5~\textsubscript{4.9}} & \cellcolor{green!20}{+5.2~\textsubscript{2.5}} & \cellcolor{green!20}{+3.6~\textsubscript{2.5}} & -0.6~\textsubscript{0.4} & \cellcolor{green!20}{+0.4~\textsubscript{2.2}} & -1.3~\textsubscript{0.6} & -0.5~\textsubscript{1.3} & \cellcolor{green!20}{+1.2~\textsubscript{0.7}} & -1.8~\textsubscript{0.5} & -0.8~\textsubscript{0.6} & \cellcolor{green!20}{+1.2~\textsubscript{0.8}} & \cellcolor{green!20}{+0.2~\textsubscript{0.6}} \\
\bottomrule
\end{tabular}}
\end{center}
\end{table*}

\begin{table*}[!t]
\small
\caption{Verbatim retrieval performance across random seeds (3B Marin). \colorbox{gray!20}{Gray} rows give the PT-Only baseline mean over checkpoints at 5K--10K steps; the rows below give the mean paired difference ($\Delta$) from that baseline, with checkpoint standard deviations as subscripts. Lower is better ($\Delta < 0$ is \colorbox{green!20}{green}).}
\label{tab:appendix_section5_verbatim_multiseed_3b}
\renewcommand{\arraystretch}{0.9}
\begin{center}
\begin{tabular}{llr}
\toprule
 & \textbf{Approach} & \textbf{Mean NLL} $\downarrow$\\
\midrule
\multirow{2}{*}{\textbf{Seed 1}} & \cellcolor{gray!20}Marin 3B (PT-Only) & \cellcolor{gray!20}3.150 \\
 & $k$-Shuffle Dyck & \cellcolor{green!20}{-0.073~\textsubscript{0.024}} \\
\midrule
\multirow{2}{*}{\textbf{Seed 2}} & \cellcolor{gray!20}Marin 3B (PT-Only) & \cellcolor{gray!20}3.101 \\
 & $k$-Shuffle Dyck & +0.014~\textsubscript{0.022} \\
\midrule
\multirow{2}{*}{\textbf{Seed 3}} & \cellcolor{gray!20}Marin 3B (PT-Only) & \cellcolor{gray!20}3.113 \\
 & $k$-Shuffle Dyck & \cellcolor{green!20}{-0.040~\textsubscript{0.044}} \\
\midrule
\multirow{2}{*}{\textbf{Seed Mean}} & \cellcolor{gray!20}Marin 3B (PT-Only) & \cellcolor{gray!20}3.121~\textsubscript{0.025} \\
 & $k$-Shuffle Dyck & \cellcolor{green!20}{-0.033~\textsubscript{0.044}} \\
\bottomrule
\end{tabular}
\end{center}
\end{table*}

Our main experiments report single runs per configuration due to compute constraints.
For instance, the 3B PT experiments alone consume approximately 1.07T tokens, which precludes repeating every cell of the grid.
Nonetheless, we quantify seed sensitivity on the configuration our conclusions rely on most: Marin at 3B.
It yields the highest absolute scores for PT-Only and the largest 3B gain for $k$-Shuffle Dyck.
We train two additional seeds per approach, giving three seeds each.
Each additional seed resamples the parameter initialization and the training data order while holding every architectural and optimization hyperparameter of Table~\ref{tab:model_hyperparameters} fixed, isolating stochastic variation from any change of setting.

Table~\ref{tab:appendix_section5_downstream_multiseed_3b} reports per-seed scores and the paired difference within each seed.
The PT-Only downstream average varies by 0.5 points across seeds (54.3, 54.0, and 54.5), and the gain from $k$-Shuffle Dyck varies by 0.2 points ($+1.9$, $+1.9$, and $+1.7$), which falls within the 0.5 checkpoint standard deviation reported for this configuration (Table~\ref{tab:section6_ablations}).
Every downstream task category retains its sign in all three seeds, and HellaSwag and LAMBADA retain their large gains across seeds.
The downstream conclusion is therefore not an artifact of a single initialization.

The two linguistic competence benchmarks behave as \S\ref{subsec:linguistic} predicts.
BLiMP overall (Table~\ref{tab:appendix_section5_blimp_multiseed_3b}) moves from $-0.3$ to $+0.9$ across seeds, with a baseline spread of 1.6 points, reproducing within one configuration the sign instability we observe across mixtures and scales.
Verbatim retrieval (Table~\ref{tab:appendix_section5_verbatim_multiseed_3b}) improves in two of the three seeds, averaging $-0.033$ NLL, with the remaining seed at $+0.014$.
The magnitude of this effect at a single configuration is thus comparable to checkpoint variance, and the retrieval account in \S\ref{sec:analysis} rests on the stable verbatim gains across task-mixture pairs rather than on this configuration alone.
Seed variation therefore preserves the direction and magnitude of the PPT effect in this setting, which is consistent with the reduced gradient noise expected under large-batch optimization (\S\ref{subsec:duration}).

\section{Supplementary Analysis}
\label{app:suppl_analysis}

\begin{table*}[t]
\begin{center}
\small
\caption{Pre-pretraining task ablation at 3B, per data mixture. Entries are mean paired differences from PT-Only over checkpoints at 5K--10K steps, with standard deviations across checkpoints as subscripts. Higher is better except for verbatim NLL.}
\label{tab:appendix_ppt_tasks_per_mixture}
\resizebox{\textwidth}{!}{
\begin{tabular}{llccccccc}
\toprule
 & & \multicolumn{5}{c}{\textbf{Downstream (\%)}} & \textbf{BLiMP} & \textbf{Verbatim} \\
\cmidrule(lr){3-7}\cmidrule(lr){8-8}\cmidrule(lr){9-9}
 & \textbf{Approach} & RC & Sci. QA & CR & LM & \textbf{Avg} & (\%) & NLL $\downarrow$ \\
\midrule
\multirow{5}{*}{\rotatebox{90}{C4}} & PT-Only (abs.) & 53.1 & 45.9 & 57.2 & 36.7 & 48.2 & 78.3 & 3.227 \\
 & $k$-Shuffle Dyck & \cellcolor{green!20}{+1.6$_{0.5}$} & \cellcolor{green!20}{+0.4$_{0.6}$} & \cellcolor{green!20}{+0.5$_{0.7}$} & \cellcolor{green!20}{+2.8$_{1.1}$} & \cellcolor{green!20}{+1.3$_{0.6}$} & \cellcolor{green!20}{+1.7$_{1.8}$} & \cellcolor{green!20}{-0.093$_{0.028}$} \\
 & MP-Struct Core & \cellcolor{green!20}{+1.1$_{0.7}$} & \cellcolor{green!20}{+1.3$_{1.4}$} & \cellcolor{green!20}{+1.1$_{1.0}$} & \cellcolor{green!20}{+3.2$_{0.7}$} & \cellcolor{green!20}{+1.7$_{0.6}$} & \cellcolor{green!20}{+2.3$_{1.9}$} & \cellcolor{green!20}{-0.077$_{0.026}$} \\
 & NCA & \cellcolor{green!20}{+0.5$_{0.3}$} & \cellcolor{green!20}{+1.9$_{0.7}$} & \cellcolor{green!20}{+0.8$_{1.1}$} & \cellcolor{green!20}{+1.4$_{0.9}$} & \cellcolor{green!20}{+1.2$_{0.4}$} & \cellcolor{green!20}{+2.1$_{1.8}$} & \cellcolor{green!20}{-0.101$_{0.033}$} \\
 & Set & -8.3$_{4.6}$ & -6.4$_{3.8}$ & -7.2$_{4.4}$ & -14.0$_{11.4}$ & -9.0$_{6.0}$ & -4.4$_{10.8}$ & +0.527$_{0.469}$ \\
\midrule
\multirow{5}{*}{\rotatebox{90}{SmolLM3}} & PT-Only (abs.) & 51.5 & 56.8 & 56.0 & 44.5 & 52.2 & 80.3 & 3.183 \\
 & $k$-Shuffle Dyck & \cellcolor{green!20}{+2.1$_{0.5}$} & \cellcolor{green!20}{+2.0$_{1.0}$} & \cellcolor{green!20}{+1.8$_{0.6}$} & \cellcolor{green!20}{+1.2$_{1.0}$} & \cellcolor{green!20}{+1.8$_{0.3}$} & \cellcolor{green!20}{+0.7$_{1.9}$} & \cellcolor{green!20}{-0.030$_{0.046}$} \\
 & MP-Struct Core & \cellcolor{green!20}{+1.1$_{0.6}$} & \cellcolor{green!20}{+1.2$_{1.0}$} & \cellcolor{green!20}{+1.3$_{0.6}$} & \cellcolor{green!20}{+0.5$_{0.7}$} & \cellcolor{green!20}{+1.0$_{0.2}$} & \cellcolor{green!20}{+1.8$_{1.5}$} & \cellcolor{green!20}{-0.019$_{0.026}$} \\
 & NCA & \cellcolor{green!20}{+1.5$_{0.9}$} & \cellcolor{green!20}{+2.4$_{0.9}$} & \cellcolor{green!20}{+1.8$_{0.5}$} & \cellcolor{green!20}{+1.5$_{1.4}$} & \cellcolor{green!20}{+1.8$_{0.3}$} & \cellcolor{green!20}{+1.2$_{2.3}$} & \cellcolor{green!20}{-0.059$_{0.030}$} \\
 & Set & -6.3$_{9.0}$ & -7.7$_{9.8}$ & -5.3$_{4.5}$ & -12.8$_{14.0}$ & -8.0$_{9.3}$ & -1.7$_{6.5}$ & +0.479$_{0.896}$ \\
\midrule
\multirow{5}{*}{\rotatebox{90}{OLMo3}} & PT-Only (abs.) & 44.0 & 41.4 & 52.1 & 33.7 & 42.8 & 77.6 & 3.673 \\
 & $k$-Shuffle Dyck & \cellcolor{green!20}{+0.4$_{0.5}$} & \cellcolor{green!20}{+0.5$_{1.4}$} & -0.5$_{0.9}$ & -0.2$_{2.9}$ & +0.0$_{1.2}$ & \cellcolor{green!20}{+0.1$_{1.7}$} & \cellcolor{green!20}{-0.056$_{0.071}$} \\
 & MP-Struct Core & \cellcolor{green!20}{+0.5$_{0.7}$} & -0.5$_{0.9}$ & -0.4$_{1.3}$ & -0.5$_{1.7}$ & -0.2$_{0.7}$ & -2.2$_{1.7}$ & \cellcolor{green!20}{-0.042$_{0.083}$} \\
 & NCA & \cellcolor{green!20}{+0.7$_{0.9}$} & -0.2$_{1.5}$ & -0.5$_{0.9}$ & -0.3$_{2.2}$ & -0.1$_{0.9}$ & \cellcolor{green!20}{+0.8$_{1.2}$} & \cellcolor{green!20}{-0.093$_{0.084}$} \\
 & Set & -3.4$_{7.9}$ & -3.6$_{5.9}$ & -2.4$_{3.0}$ & -6.2$_{12.5}$ & -3.9$_{7.3}$ & -3.3$_{6.5}$ & +0.371$_{0.972}$ \\
\midrule
\multirow{5}{*}{\rotatebox{90}{Marin}} & PT-Only (abs.) & 54.4 & 56.1 & 58.4 & 48.5 & 54.3 & 79.7 & 3.150 \\
 & $k$-Shuffle Dyck & \cellcolor{green!20}{+0.6$_{0.6}$} & \cellcolor{green!20}{+2.3$_{0.6}$} & \cellcolor{green!20}{+1.8$_{0.8}$} & \cellcolor{green!20}{+3.1$_{1.3}$} & \cellcolor{green!20}{+1.9$_{0.5}$} & -0.3$_{1.0}$ & \cellcolor{green!20}{-0.073$_{0.024}$} \\
 & MP-Struct Core & +0.0$_{0.6}$ & \cellcolor{green!20}{+1.4$_{0.8}$} & \cellcolor{green!20}{+0.7$_{1.1}$} & \cellcolor{green!20}{+2.9$_{0.9}$} & \cellcolor{green!20}{+1.3$_{0.3}$} & \cellcolor{green!20}{+0.6$_{2.0}$} & \cellcolor{green!20}{-0.039$_{0.029}$} \\
 & NCA & \cellcolor{green!20}{+0.5$_{0.5}$} & \cellcolor{green!20}{+2.2$_{1.0}$} & \cellcolor{green!20}{+1.1$_{0.3}$} & \cellcolor{green!20}{+3.1$_{1.6}$} & \cellcolor{green!20}{+1.7$_{0.6}$} & \cellcolor{green!20}{+1.2$_{2.0}$} & \cellcolor{green!20}{-0.075$_{0.020}$} \\
 & Set & -6.8$_{4.9}$ & -7.0$_{5.4}$ & -5.9$_{4.2}$ & -11.6$_{12.5}$ & -7.8$_{6.7}$ & -0.0$_{6.1}$ & +0.973$_{1.300}$ \\
\bottomrule
\end{tabular}
}
\end{center}
\end{table*}

\begin{table*}[t]
\small
\caption{Absolute scores for the extended-budget runs. Baselines (\colorbox{gray!20}{PT-Only}) are shaded gray; \colorbox{green!20}{green} and \textbf{bold} mark improvements from PPT ($k$-Shuffle Dyck).}
\begin{center}
\label{tab:appendix_scaling_absolute}
\resizebox{\textwidth}{!}{
\begin{tabular}{lllccccccc}
\toprule
\textbf{Setting} & \textbf{Tokens} & \textbf{Approach} & \textbf{RC} & \textbf{Science QA} & \textbf{CR} & \textbf{LM} & \textbf{Avg (\%)} & \textbf{BLiMP (\%)} & \textbf{Verbatim NLL $\downarrow$} \\
\midrule
\multirow{10}{*}{\textbf{C4 (3B)}}
 & \multirow{2}{*}{21B} & \cellcolor{gray!20}PT-Only & \cellcolor{gray!20}55.0 & \cellcolor{gray!20}46.0 & \cellcolor{gray!20}59.0 & \cellcolor{gray!20}39.0 & \cellcolor{gray!20}49.7 & \cellcolor{gray!20}77.8 & \cellcolor{gray!20}3.165 \\
 &  & PPT ($k$-Shuffle Dyck) & \cellcolor{green!20}\textbf{55.5} & \cellcolor{green!20}\textbf{47.8} & \cellcolor{green!20}\textbf{59.4} & \cellcolor{green!20}\textbf{41.2} & \cellcolor{green!20}\textbf{51.0} & \cellcolor{green!20}\textbf{80.8} & \cellcolor{green!20}\textbf{3.127} \\
\cmidrule(lr){2-10}
 & \multirow{2}{*}{42B} & \cellcolor{gray!20}PT-Only & \cellcolor{gray!20}57.0 & \cellcolor{gray!20}51.0 & \cellcolor{gray!20}60.5 & \cellcolor{gray!20}44.2 & \cellcolor{gray!20}53.2 & \cellcolor{gray!20}77.9 & \cellcolor{gray!20}3.062 \\
 &  & PPT ($k$-Shuffle Dyck) & \cellcolor{green!20}\textbf{58.5} & 49.2 & \cellcolor{green!20}\textbf{61.8} & \cellcolor{green!20}\textbf{46.0} & \cellcolor{green!20}\textbf{53.9} & \cellcolor{green!20}\textbf{81.0} & 3.077 \\
\cmidrule(lr){2-10}
 & \multirow{2}{*}{63B} & \cellcolor{gray!20}PT-Only & \cellcolor{gray!20}58.3 & \cellcolor{gray!20}53.0 & \cellcolor{gray!20}61.8 & \cellcolor{gray!20}44.6 & \cellcolor{gray!20}54.4 & \cellcolor{gray!20}78.2 & \cellcolor{gray!20}3.057 \\
 &  & PPT ($k$-Shuffle Dyck) & \cellcolor{green!20}\textbf{58.8} & 52.9 & \cellcolor{green!20}\textbf{63.0} & \cellcolor{green!20}\textbf{46.1} & \cellcolor{green!20}\textbf{55.2} & 75.9 & \cellcolor{green!20}\textbf{3.025} \\
\cmidrule(lr){2-10}
 & \multirow{2}{*}{84B} & \cellcolor{gray!20}PT-Only & \cellcolor{gray!20}59.0 & \cellcolor{gray!20}55.0 & \cellcolor{gray!20}63.7 & \cellcolor{gray!20}46.1 & \cellcolor{gray!20}55.9 & \cellcolor{gray!20}80.2 & \cellcolor{gray!20}3.062 \\
 &  & PPT ($k$-Shuffle Dyck) & \cellcolor{green!20}\textbf{59.6} & 54.2 & \cellcolor{green!20}\textbf{64.5} & \cellcolor{green!20}\textbf{48.8} & \cellcolor{green!20}\textbf{56.8} & 79.5 & \cellcolor{green!20}\textbf{3.041} \\
\cmidrule(lr){2-10}
 & \multirow{2}{*}{100B} & \cellcolor{gray!20}PT-Only & \cellcolor{gray!20}60.1 & \cellcolor{gray!20}56.0 & \cellcolor{gray!20}65.0 & \cellcolor{gray!20}48.7 & \cellcolor{gray!20}57.5 & \cellcolor{gray!20}79.9 & \cellcolor{gray!20}2.958 \\
 &  & PPT ($k$-Shuffle Dyck) & \cellcolor{green!20}\textbf{61.3} & 56.0 & \cellcolor{green!20}\textbf{66.2} & \cellcolor{green!20}\textbf{52.7} & \cellcolor{green!20}\textbf{59.1} & 64.4 & 2.966 \\
\midrule
\multirow{10}{*}{\textbf{Marin (3B)}}
 & \multirow{2}{*}{21B} & \cellcolor{gray!20}PT-Only & \cellcolor{gray!20}55.7 & \cellcolor{gray!20}58.2 & \cellcolor{gray!20}59.3 & \cellcolor{gray!20}50.9 & \cellcolor{gray!20}56.0 & \cellcolor{gray!20}82.3 & \cellcolor{gray!20}3.121 \\
 &  & PPT ($k$-Shuffle Dyck) & \cellcolor{green!20}\textbf{56.0} & \cellcolor{green!20}\textbf{59.3} & \cellcolor{green!20}\textbf{60.4} & \cellcolor{green!20}\textbf{52.7} & \cellcolor{green!20}\textbf{57.1} & 79.2 & \cellcolor{green!20}\textbf{3.106} \\
\cmidrule(lr){2-10}
 & \multirow{2}{*}{42B} & \cellcolor{gray!20}PT-Only & \cellcolor{gray!20}58.1 & \cellcolor{gray!20}62.1 & \cellcolor{gray!20}62.7 & \cellcolor{gray!20}55.7 & \cellcolor{gray!20}59.7 & \cellcolor{gray!20}82.5 & \cellcolor{gray!20}3.020 \\
 &  & PPT ($k$-Shuffle Dyck) & \cellcolor{green!20}\textbf{58.5} & \cellcolor{green!20}\textbf{64.0} & \cellcolor{green!20}\textbf{64.3} & \cellcolor{green!20}\textbf{58.5} & \cellcolor{green!20}\textbf{61.3} & 72.2 & \cellcolor{green!20}\textbf{3.006} \\
\cmidrule(lr){2-10}
 & \multirow{2}{*}{63B} & \cellcolor{gray!20}PT-Only & \cellcolor{gray!20}58.2 & \cellcolor{gray!20}62.8 & \cellcolor{gray!20}63.5 & \cellcolor{gray!20}58.6 & \cellcolor{gray!20}60.8 & \cellcolor{gray!20}81.8 & \cellcolor{gray!20}2.959 \\
 &  & PPT ($k$-Shuffle Dyck) & \cellcolor{green!20}\textbf{59.1} & \cellcolor{green!20}\textbf{64.5} & \cellcolor{green!20}\textbf{64.4} & \cellcolor{green!20}\textbf{61.1} & \cellcolor{green!20}\textbf{62.3} & 80.7 & \cellcolor{green!20}\textbf{2.940} \\
\cmidrule(lr){2-10}
 & \multirow{2}{*}{84B} & \cellcolor{gray!20}PT-Only & \cellcolor{gray!20}60.0 & \cellcolor{gray!20}64.3 & \cellcolor{gray!20}64.9 & \cellcolor{gray!20}58.7 & \cellcolor{gray!20}62.0 & \cellcolor{gray!20}81.1 & \cellcolor{gray!20}2.984 \\
 &  & PPT ($k$-Shuffle Dyck) & \cellcolor{green!20}\textbf{60.3} & \cellcolor{green!20}\textbf{65.7} & \cellcolor{green!20}\textbf{65.1} & \cellcolor{green!20}\textbf{62.0} & \cellcolor{green!20}\textbf{63.3} & \cellcolor{green!20}\textbf{82.4} & \cellcolor{green!20}\textbf{2.949} \\
\cmidrule(lr){2-10}
 & \multirow{2}{*}{100B} & \cellcolor{gray!20}PT-Only & \cellcolor{gray!20}61.5 & \cellcolor{gray!20}66.7 & \cellcolor{gray!20}67.4 & \cellcolor{gray!20}63.3 & \cellcolor{gray!20}64.7 & \cellcolor{gray!20}83.0 & \cellcolor{gray!20}2.962 \\
 &  & PPT ($k$-Shuffle Dyck) & 61.2 & \cellcolor{green!20}\textbf{68.0} & \cellcolor{green!20}\textbf{68.9} & \cellcolor{green!20}\textbf{65.5} & \cellcolor{green!20}\textbf{65.9} & 82.5 & \cellcolor{green!20}\textbf{2.923} \\
\midrule
\multirow{8}{*}{\textbf{Marin (7B)}}
 & \multirow{2}{*}{21B} & \cellcolor{gray!20}PT-Only & \cellcolor{gray!20}55.4 & \cellcolor{gray!20}60.4 & \cellcolor{gray!20}61.0 & \cellcolor{gray!20}51.8 & \cellcolor{gray!20}57.1 & \cellcolor{gray!20}81.2 & \cellcolor{gray!20}3.104 \\
 &  & PPT ($k$-Shuffle Dyck) & \cellcolor{green!20}\textbf{56.4} & 60.2 & \cellcolor{green!20}\textbf{61.5} & \cellcolor{green!20}\textbf{52.6} & \cellcolor{green!20}\textbf{57.6} & 81.0 & 3.113 \\
\cmidrule(lr){2-10}
 & \multirow{2}{*}{42B} & \cellcolor{gray!20}PT-Only & \cellcolor{gray!20}59.3 & \cellcolor{gray!20}63.9 & \cellcolor{gray!20}64.9 & \cellcolor{gray!20}57.7 & \cellcolor{gray!20}61.4 & \cellcolor{gray!20}80.4 & \cellcolor{gray!20}2.957 \\
 &  & PPT ($k$-Shuffle Dyck) & 59.0 & 63.8 & \cellcolor{green!20}\textbf{65.0} & \cellcolor{green!20}\textbf{59.2} & \cellcolor{green!20}\textbf{61.8} & \cellcolor{green!20}\textbf{82.2} & 2.959 \\
\cmidrule(lr){2-10}
 & \multirow{2}{*}{63B} & \cellcolor{gray!20}PT-Only & \cellcolor{gray!20}59.1 & \cellcolor{gray!20}65.1 & \cellcolor{gray!20}66.2 & \cellcolor{gray!20}61.6 & \cellcolor{gray!20}63.0 & \cellcolor{gray!20}82.0 & \cellcolor{gray!20}2.925 \\
 &  & PPT ($k$-Shuffle Dyck) & \cellcolor{green!20}\textbf{60.0} & \cellcolor{green!20}\textbf{67.1} & 65.9 & 60.8 & \cellcolor{green!20}\textbf{63.5} & 81.4 & \cellcolor{green!20}\textbf{2.921} \\
\cmidrule(lr){2-10}
 & \multirow{2}{*}{75.5B} & \cellcolor{gray!20}PT-Only & \cellcolor{gray!20}60.0 & \cellcolor{gray!20}66.2 & \cellcolor{gray!20}66.6 & \cellcolor{gray!20}63.2 & \cellcolor{gray!20}64.0 & \cellcolor{gray!20}82.9 & \cellcolor{gray!20}2.934 \\
 &  & PPT ($k$-Shuffle Dyck) & \cellcolor{green!20}\textbf{60.3} & \cellcolor{green!20}\textbf{67.7} & \cellcolor{green!20}\textbf{68.4} & \cellcolor{green!20}\textbf{63.6} & \cellcolor{green!20}\textbf{65.0} & 81.8 & \cellcolor{green!20}\textbf{2.905} \\
\bottomrule
\end{tabular}
}
\end{center}
\end{table*}

\begin{table*}[t]
\small
\caption{Task-level breakdown for the extended-budget runs. Baselines (\colorbox{gray!20}{PT-Only}) are shaded gray; \colorbox{green!20}{green} and \textbf{bold} mark improvements from PPT ($k$-Shuffle Dyck).}
\label{tab:appendix_scaling_tasks}
\renewcommand{\arraystretch}{1.0}
\setlength{\tabcolsep}{3pt}
\begin{center}
\resizebox{\textwidth}{!}{
\begin{tabular}{lllcccccccccc}
\toprule
& & & \multicolumn{2}{c}{\textbf{RC}} & \multicolumn{3}{c}{\textbf{Science QA}} & \multicolumn{4}{c}{\textbf{CR}} & \textbf{LM} \\
\cmidrule(lr){4-5} \cmidrule(lr){6-8} \cmidrule(lr){9-12} \cmidrule(lr){13-13}
\textbf{Setting} & \textbf{Tokens} & \textbf{Approach} & \textbf{RACE} & \textbf{ReCoRD} & \textbf{SciQ} & \textbf{ARC-Easy} & \textbf{OpenBookQA} & \textbf{COPA} & \textbf{PIQA} & \textbf{SIQA} & \textbf{HellaSwag} & \textbf{LAMBADA} \\
\midrule
\multirow{10}{*}{\textbf{C4 (3B)}}
 & \multirow{2}{*}{21B} & \cellcolor{gray!20}PT-Only & \cellcolor{gray!20}33.2 & \cellcolor{gray!20}76.8 & \cellcolor{gray!20}67.8 & \cellcolor{gray!20}41.3 & \cellcolor{gray!20}28.8 & \cellcolor{gray!20}76.0 & \cellcolor{gray!20}71.7 & \cellcolor{gray!20}36.9 & \cellcolor{gray!20}51.2 & \cellcolor{gray!20}39.0 \\
 &  & PPT & 33.2 & \cellcolor{green!20}\textbf{77.8} & \cellcolor{green!20}\textbf{69.5} & 39.8 & \cellcolor{green!20}\textbf{34.0} & 75.0 & \cellcolor{green!20}\textbf{73.2} & 35.9 & \cellcolor{green!20}\textbf{53.7} & \cellcolor{green!20}\textbf{41.2} \\
\cmidrule(lr){2-13}
 & \multirow{2}{*}{42B} & \cellcolor{gray!20}PT-Only & \cellcolor{gray!20}34.3 & \cellcolor{gray!20}79.7 & \cellcolor{gray!20}72.9 & \cellcolor{gray!20}49.0 & \cellcolor{gray!20}31.0 & \cellcolor{gray!20}75.0 & \cellcolor{gray!20}72.6 & \cellcolor{gray!20}37.7 & \cellcolor{gray!20}56.6 & \cellcolor{gray!20}44.2 \\
 &  & PPT & \cellcolor{green!20}\textbf{36.0} & \cellcolor{green!20}\textbf{81.0} & \cellcolor{green!20}\textbf{74.3} & 40.1 & \cellcolor{green!20}\textbf{33.2} & \cellcolor{green!20}\textbf{76.0} & \cellcolor{green!20}\textbf{75.0} & 37.4 & \cellcolor{green!20}\textbf{58.8} & \cellcolor{green!20}\textbf{46.0} \\
\cmidrule(lr){2-13}
 & \multirow{2}{*}{63B} & \cellcolor{gray!20}PT-Only & \cellcolor{gray!20}35.9 & \cellcolor{gray!20}80.8 & \cellcolor{gray!20}77.2 & \cellcolor{gray!20}50.8 & \cellcolor{gray!20}31.0 & \cellcolor{gray!20}73.0 & \cellcolor{gray!20}74.9 & \cellcolor{gray!20}40.7 & \cellcolor{gray!20}58.6 & \cellcolor{gray!20}44.6 \\
 &  & PPT & 35.2 & \cellcolor{green!20}\textbf{82.4} & \cellcolor{green!20}\textbf{78.6} & 46.0 & \cellcolor{green!20}\textbf{34.0} & \cellcolor{green!20}\textbf{77.0} & 74.9 & 38.8 & \cellcolor{green!20}\textbf{61.5} & \cellcolor{green!20}\textbf{46.1} \\
\cmidrule(lr){2-13}
 & \multirow{2}{*}{84B} & \cellcolor{gray!20}PT-Only & \cellcolor{gray!20}35.9 & \cellcolor{gray!20}82.2 & \cellcolor{gray!20}76.5 & \cellcolor{gray!20}53.5 & \cellcolor{gray!20}35.0 & \cellcolor{gray!20}77.0 & \cellcolor{gray!20}74.4 & \cellcolor{gray!20}43.0 & \cellcolor{gray!20}60.3 & \cellcolor{gray!20}46.1 \\
 &  & PPT & \cellcolor{green!20}\textbf{36.4} & \cellcolor{green!20}\textbf{82.9} & \cellcolor{green!20}\textbf{78.2} & 50.6 & 33.8 & \cellcolor{green!20}\textbf{79.0} & \cellcolor{green!20}\textbf{76.0} & 40.5 & \cellcolor{green!20}\textbf{62.4} & \cellcolor{green!20}\textbf{48.8} \\
\cmidrule(lr){2-13}
 & \multirow{2}{*}{100B} & \cellcolor{gray!20}PT-Only & \cellcolor{gray!20}36.2 & \cellcolor{gray!20}84.0 & \cellcolor{gray!20}76.2 & \cellcolor{gray!20}56.9 & \cellcolor{gray!20}35.0 & \cellcolor{gray!20}78.0 & \cellcolor{gray!20}76.3 & \cellcolor{gray!20}42.0 & \cellcolor{gray!20}63.8 & \cellcolor{gray!20}48.7 \\
 &  & PPT & \cellcolor{green!20}\textbf{37.2} & \cellcolor{green!20}\textbf{85.4} & \cellcolor{green!20}\textbf{80.1} & 53.9 & 34.0 & \cellcolor{green!20}\textbf{81.0} & \cellcolor{green!20}\textbf{77.1} & 40.3 & \cellcolor{green!20}\textbf{66.3} & \cellcolor{green!20}\textbf{52.7} \\
\midrule
\multirow{10}{*}{\textbf{Marin (3B)}}
 & \multirow{2}{*}{21B} & \cellcolor{gray!20}PT-Only & \cellcolor{gray!20}34.0 & \cellcolor{gray!20}77.5 & \cellcolor{gray!20}77.4 & \cellcolor{gray!20}63.0 & \cellcolor{gray!20}34.2 & \cellcolor{gray!20}71.0 & \cellcolor{gray!20}71.5 & \cellcolor{gray!20}43.5 & \cellcolor{gray!20}51.2 & \cellcolor{gray!20}50.9 \\
 &  & PPT & 33.5 & \cellcolor{green!20}\textbf{78.5} & \cellcolor{green!20}\textbf{79.6} & \cellcolor{green!20}\textbf{64.2} & 34.2 & \cellcolor{green!20}\textbf{72.0} & \cellcolor{green!20}\textbf{72.4} & \cellcolor{green!20}\textbf{44.0} & \cellcolor{green!20}\textbf{53.5} & \cellcolor{green!20}\textbf{52.7} \\
\cmidrule(lr){2-13}
 & \multirow{2}{*}{42B} & \cellcolor{gray!20}PT-Only & \cellcolor{gray!20}35.4 & \cellcolor{gray!20}80.7 & \cellcolor{gray!20}82.4 & \cellcolor{gray!20}65.9 & \cellcolor{gray!20}38.0 & \cellcolor{gray!20}77.0 & \cellcolor{gray!20}72.8 & \cellcolor{gray!20}44.4 & \cellcolor{gray!20}56.7 & \cellcolor{gray!20}55.7 \\
 &  & PPT & 35.3 & \cellcolor{green!20}\textbf{81.7} & \cellcolor{green!20}\textbf{85.4} & \cellcolor{green!20}\textbf{67.7} & \cellcolor{green!20}\textbf{38.8} & \cellcolor{green!20}\textbf{79.0} & \cellcolor{green!20}\textbf{74.2} & \cellcolor{green!20}\textbf{45.2} & \cellcolor{green!20}\textbf{58.7} & \cellcolor{green!20}\textbf{58.5} \\
\cmidrule(lr){2-13}
 & \multirow{2}{*}{63B} & \cellcolor{gray!20}PT-Only & \cellcolor{gray!20}34.6 & \cellcolor{gray!20}81.8 & \cellcolor{gray!20}84.5 & \cellcolor{gray!20}68.7 & \cellcolor{gray!20}35.2 & \cellcolor{gray!20}75.0 & \cellcolor{gray!20}73.9 & \cellcolor{gray!20}45.0 & \cellcolor{gray!20}60.0 & \cellcolor{gray!20}58.6 \\
 &  & PPT & \cellcolor{green!20}\textbf{34.9} & \cellcolor{green!20}\textbf{83.2} & \cellcolor{green!20}\textbf{87.4} & \cellcolor{green!20}\textbf{69.7} & \cellcolor{green!20}\textbf{36.4} & \cellcolor{green!20}\textbf{77.0} & \cellcolor{green!20}\textbf{74.6} & 45.0 & \cellcolor{green!20}\textbf{61.1} & \cellcolor{green!20}\textbf{61.1} \\
\cmidrule(lr){2-13}
 & \multirow{2}{*}{84B} & \cellcolor{gray!20}PT-Only & \cellcolor{gray!20}36.6 & \cellcolor{gray!20}83.4 & \cellcolor{gray!20}87.1 & \cellcolor{gray!20}69.1 & \cellcolor{gray!20}36.8 & \cellcolor{gray!20}79.0 & \cellcolor{gray!20}74.3 & \cellcolor{gray!20}45.5 & \cellcolor{gray!20}60.8 & \cellcolor{gray!20}58.7 \\
 &  & PPT & 36.5 & \cellcolor{green!20}\textbf{84.2} & \cellcolor{green!20}\textbf{88.6} & \cellcolor{green!20}\textbf{71.0} & \cellcolor{green!20}\textbf{37.4} & 77.0 & \cellcolor{green!20}\textbf{74.5} & \cellcolor{green!20}\textbf{46.7} & \cellcolor{green!20}\textbf{62.0} & \cellcolor{green!20}\textbf{62.0} \\
\cmidrule(lr){2-13}
 & \multirow{2}{*}{100B} & \cellcolor{gray!20}PT-Only & \cellcolor{gray!20}37.7 & \cellcolor{gray!20}85.2 & \cellcolor{gray!20}88.7 & \cellcolor{gray!20}72.1 & \cellcolor{gray!20}39.2 & \cellcolor{gray!20}83.0 & \cellcolor{gray!20}75.5 & \cellcolor{gray!20}46.1 & \cellcolor{gray!20}64.9 & \cellcolor{gray!20}63.3 \\
 &  & PPT & 36.7 & \cellcolor{green!20}\textbf{85.8} & \cellcolor{green!20}\textbf{89.5} & \cellcolor{green!20}\textbf{74.3} & \cellcolor{green!20}\textbf{40.2} & \cellcolor{green!20}\textbf{85.0} & \cellcolor{green!20}\textbf{76.7} & \cellcolor{green!20}\textbf{47.7} & \cellcolor{green!20}\textbf{66.1} & \cellcolor{green!20}\textbf{65.5} \\
\midrule
\multirow{8}{*}{\textbf{Marin (7B)}}
 & \multirow{2}{*}{21B} & \cellcolor{gray!20}PT-Only & \cellcolor{gray!20}32.8 & \cellcolor{gray!20}78.0 & \cellcolor{gray!20}78.9 & \cellcolor{gray!20}65.7 & \cellcolor{gray!20}36.6 & \cellcolor{gray!20}74.0 & \cellcolor{gray!20}72.2 & \cellcolor{gray!20}44.0 & \cellcolor{gray!20}53.8 & \cellcolor{gray!20}51.8 \\
 &  & PPT & \cellcolor{green!20}\textbf{34.9} & 77.8 & 78.9 & 65.2 & 36.4 & \cellcolor{green!20}\textbf{76.0} & \cellcolor{green!20}\textbf{72.5} & 43.2 & \cellcolor{green!20}\textbf{54.2} & \cellcolor{green!20}\textbf{52.6} \\
\cmidrule(lr){2-13}
 & \multirow{2}{*}{42B} & \cellcolor{gray!20}PT-Only & \cellcolor{gray!20}36.1 & \cellcolor{gray!20}82.5 & \cellcolor{gray!20}83.5 & \cellcolor{gray!20}70.0 & \cellcolor{gray!20}38.2 & \cellcolor{gray!20}80.0 & \cellcolor{gray!20}73.3 & \cellcolor{gray!20}46.2 & \cellcolor{gray!20}60.2 & \cellcolor{gray!20}57.7 \\
 &  & PPT & 35.6 & 82.4 & \cellcolor{green!20}\textbf{86.3} & 67.4 & 37.6 & \cellcolor{green!20}\textbf{81.0} & \cellcolor{green!20}\textbf{73.6} & 44.4 & \cellcolor{green!20}\textbf{61.2} & \cellcolor{green!20}\textbf{59.2} \\
\cmidrule(lr){2-13}
 & \multirow{2}{*}{63B} & \cellcolor{gray!20}PT-Only & \cellcolor{gray!20}34.4 & \cellcolor{gray!20}83.9 & \cellcolor{gray!20}85.0 & \cellcolor{gray!20}71.7 & \cellcolor{gray!20}38.6 & \cellcolor{gray!20}81.0 & \cellcolor{gray!20}75.2 & \cellcolor{gray!20}45.1 & \cellcolor{gray!20}63.5 & \cellcolor{gray!20}61.6 \\
 &  & PPT & \cellcolor{green!20}\textbf{35.3} & \cellcolor{green!20}\textbf{84.7} & \cellcolor{green!20}\textbf{88.4} & \cellcolor{green!20}\textbf{72.4} & \cellcolor{green!20}\textbf{40.6} & 79.0 & 75.1 & \cellcolor{green!20}\textbf{45.2} & \cellcolor{green!20}\textbf{64.1} & 60.8 \\
\cmidrule(lr){2-13}
 & \multirow{2}{*}{75.5B} & \cellcolor{gray!20}PT-Only & \cellcolor{gray!20}35.5 & \cellcolor{gray!20}84.5 & \cellcolor{gray!20}88.3 & \cellcolor{gray!20}70.4 & \cellcolor{gray!20}39.8 & \cellcolor{gray!20}80.0 & \cellcolor{gray!20}74.9 & \cellcolor{gray!20}46.6 & \cellcolor{gray!20}65.1 & \cellcolor{gray!20}63.2 \\
 &  & PPT & 35.4 & \cellcolor{green!20}\textbf{85.2} & \cellcolor{green!20}\textbf{90.1} & \cellcolor{green!20}\textbf{72.3} & \cellcolor{green!20}\textbf{40.8} & \cellcolor{green!20}\textbf{86.0} & \cellcolor{green!20}\textbf{76.2} & 46.0 & \cellcolor{green!20}\textbf{65.4} & \cellcolor{green!20}\textbf{63.6} \\
\bottomrule
\end{tabular}
}
\end{center}
\end{table*}

\begin{itemize}[leftmargin=*]
\item Table~\ref{tab:appendix_ppt_tasks_per_mixture} shows the per-mixture breakdown of the PPT task ablation.
\item Table \ref{tab:appendix_scaling_absolute} shows the full absolute scores for the 100B PT runs, grouped by task category.
\item Table \ref{tab:appendix_scaling_tasks} shows the task-level breakdown for the 100B PT runs.
\end{itemize}

\end{document}